\documentclass[10pt,journal]{IEEEtran}
\ifCLASSOPTIONcompsoc
  \usepackage[nocompress]{cite}
\else
  \usepackage{cite}
\fi

\usepackage{amsmath,amssymb,amsfonts}
\usepackage{bm}
\usepackage{dsfont}
\usepackage{pifont}
\usepackage{graphicx}
\usepackage{adjustbox}
\usepackage[bookmarks=false]{hyperref}
\usepackage{comment}
\usepackage{algorithm}
\usepackage{algpseudocode}
\usepackage{units}
\usepackage{multirow}
\usepackage{array}
\usepackage{booktabs}

\usepackage{xcolor}
\definecolor{darkspringgreen}{rgb}{0.09, 0.45, 0.27}

\usepackage[normalem]{ulem}

\usepackage[normalem]{ulem}

\usepackage{tikz}
\usepackage{textcomp}
\usepackage[doipre={DOI:~}]{uri}
\usepackage{lipsum}
\newcommand\copyrighttext{%
  \scriptsize ©2025 IEEE. Personal use of this material is permitted. Permission from IEEE must be obtained for all other uses, in any current or future media, including reprinting/republishing this material for advertising or promotional purposes, creating new collective works, for resale or redistribution to servers or lists, or reuse of any copyrighted component of this work in other works. Accepted for publication in the IEEE Transactions on Computer-Aided Design of Integrated Circuits and Systems: \url{https://doi.org/10.1109/TCAD.2025.3543715}.
  }
\newcommand{\copyrightnotice}{%
\begin{tikzpicture}[remember picture,overlay]
\node[anchor=south,yshift=8pt] at (current page.south) {\fbox{\parbox{\dimexpr\textwidth-\fboxsep-\fboxrule\relax}{\copyrighttext}}};
\end{tikzpicture}%
}

\begin{document}
\bstctlcite{IEEEexample:BSTcontrol}
\title{\vspace{-1cm}Optimizing DNN Inference on Multi-Accelerator SoCs at Training-time}
\author{Matteo~Risso,~\IEEEmembership{Student Member,~IEEE,}
        Alessio~Burrello,~\IEEEmembership{Member,~IEEE,}
        and~Daniele~Jahier~Pagliari,~\IEEEmembership{Member,~IEEE}

\thanks{This work has received funding from the Key Digital Technologies Joint
Undertaking (KDT-JU) under grant agreement No 101095947. The JU receives support from the European Union’s Horizon Europe research and
innovation programme.}
\thanks{We acknowledge the CINECA award under the ISCRA initiative, for the
availability of high performance computing resources and support.}
\IEEEcompsocitemizethanks{\IEEEcompsocthanksitem M. Risso, A. Burrello and D. Jahier Pagliari are with the Department
of Control and Computer Engineering, Politecnico di Torino, 10129, Turin, Italy. E-mail: name.firstsurname@polito.it
}
\vspace{-1.5cm}
}

\IEEEtitleabstractindextext{%
\begin{abstract}
The demand for executing Deep Neural Networks (DNNs) with low latency and minimal power consumption at the edge has led to the development of advanced heterogeneous Systems-on-Chips (SoCs) that incorporate multiple specialized computing units (CUs), such as accelerators.
Offloading DNN computations to a specific CU from the available set often exposes accuracy vs efficiency trade-offs, due to differences in their supported operations (e.g., standard vs. depthwise convolution) or data representations (e.g., more/less aggressively quantized).
A challenging yet unresolved issue is how to map a DNN onto these multi-CU systems to maximally exploit the parallelization possibilities while taking accuracy into account.
To address this problem, we present ODiMO, a hardware-aware tool that efficiently explores fine-grain mapping of DNNs among various on-chip CUs, during the training phase. ODiMO strategically splits individual layers of the neural network and executes them in parallel on the multiple available CUs, aiming to balance the total inference energy consumption or latency with the resulting accuracy, impacted by the unique features of the different hardware units.
We test our approach on CIFAR-10, CIFAR-100, and ImageNet, targeting two open-source heterogeneous SoCs, i.e., DIANA and Darkside.
We obtain a rich collection of Pareto-optimal networks in the accuracy vs. energy or latency space.
We show that ODiMO reduces the latency of a DNN executed on the Darkside SoC by up to $8\times$ at iso-accuracy, compared to manual heuristic mappings. When targeting energy, on the same SoC, ODiMO produced up to $50.8\times$ more efficient mappings, with minimal accuracy drop ($< 0.3\%$).
\end{abstract}

}

\maketitle
\copyrightnotice

\IEEEdisplaynontitleabstractindextext
\IEEEpeerreviewmaketitle

\ifCLASSOPTIONcompsoc
\IEEEraisesectionheading{\section{Introduction}\label{sec:introduction}}
\else
\section{Introduction}
\label{sec:intro}
\fi
\begin{figure}[t]
  \centering
  \includegraphics[width=\columnwidth]{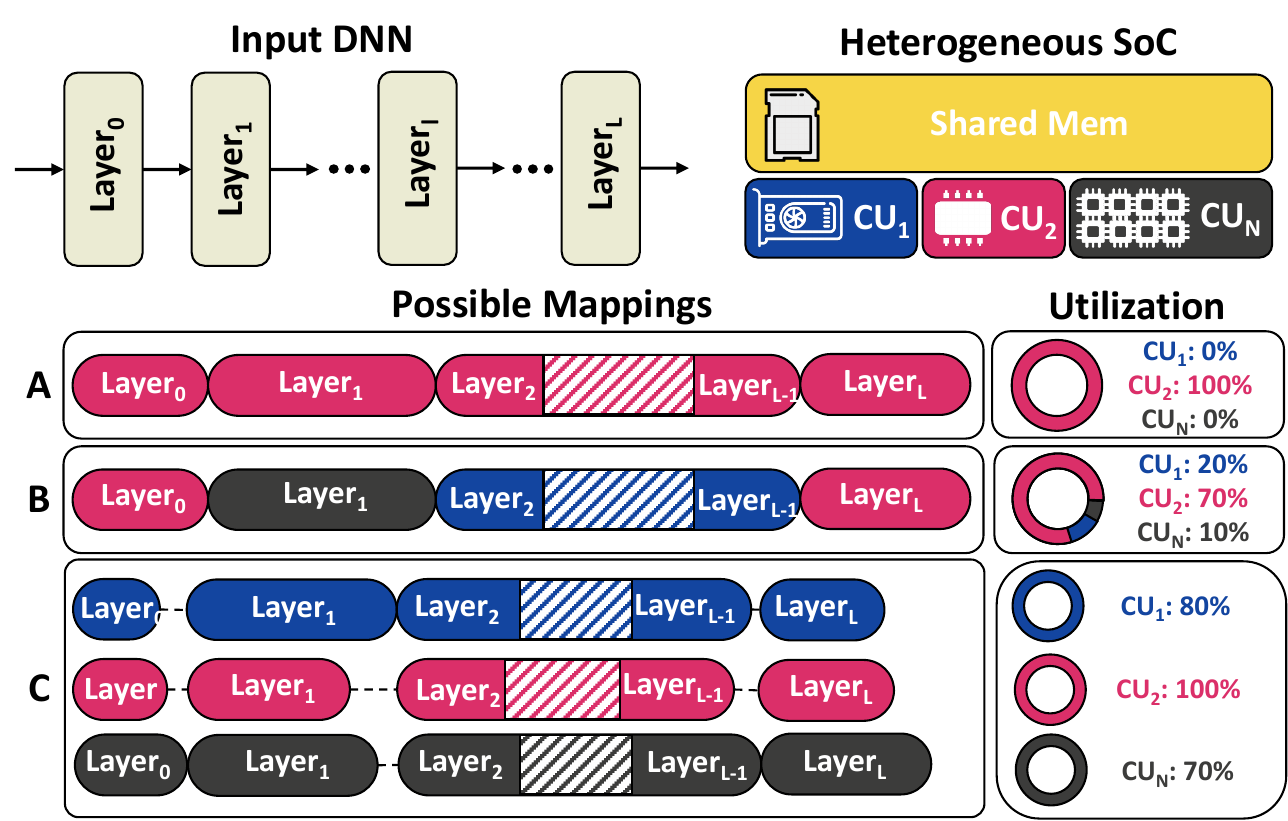}
  \caption{Different possible execution strategies of a DNN on a Heterogeneous SoC with shared memory. Strategy A offloads the DNN computation to a single CU. Strategy B uses a layer-wise scheme where each layer can be computed by one of the available CU. Strategy C is the intra-layer mapping scheme enabled by ODiMO, where the computation of each layer can be split among the different CUs.}
  \label{fig:intro}
\end{figure}
Deploying DNNs for inference at the edge offers well-known advantages in terms of latency, predictability, energy consumption, and data privacy~\cite{Zhou2019,sze2020efficient}.
However, the execution of computationally intensive DNNs on edge devices, which operate under stringent energy and memory constraints, poses a significant challenge. Current research addresses this problem in multiple, complementary ways.
On the software side, various optimization techniques are employed to enhance the efficiency and accuracy of DNN models. These include \textit{Neural Architecture Search (NAS)}, which automates the design of DNN architectures under specified resource limits, \textit{pruning}, which reduces the model size by removing redundant parameters, and \textit{quantization}, which decreases the precision of the model weights and activations to lower computational demands and memory usage~\cite{sze2020efficient, Jacob2018, pit_journal}.
On the hardware side, efficiency is mostly improved through specialization. This involves the design and integration of heterogeneous Systems-on-Chip (SoCs) that incorporate domain-specific Computing Units (CUs) dedicated to DNN processing. These specialized hardware CUs are tailored to handle specific computational patterns of DNN workloads, thereby significantly enhancing performance and/or energy efficiency~\cite{edgetpu, Dagli2022, hp_ssc2020, ueyoshi2022diana, darkside}, while sacrificing generality.

Optimizing a DNN model for execution on these multi-CU systems remains a significant challenge. 
Traditionally, entire networks are executed on a single CU, as depicted Fig.~\ref{fig:intro} (Mapping A).
More recent research has explored multi-CU inference~\cite{Wang2020, Vasiliadis2022, Tu2019, Dagli2022, Jeong2022, Kang2017} with layer-wise partitioning (as in Mapping B). 
However, layer-wise partitioning leads to suboptimal hardware utilization, since for a standard sequential DNN, only one CU is active at any time. The possibility of partitioning at a finer grain, with each layer partitioned among multiple CUs (as in Mapping C of Fig.~\ref{fig:intro}) is studied almost exclusively for the case of homogeneous CUs (e.g., multiple GPUs/TPUs)~\cite{Song2020, zheng2022alpa}, and still under-explored for systems with heterogeneous ones. 
Moreover, these studies generally assume that \textit{all CUs can execute all layers} of a DNN, and deliver equally accurate results. However, this assumption does not hold in many real-world cases~\cite{ueyoshi2022diana, hp_ssc2020, darkside}.
A notable counter-example are SoCs that incorporate both Digital and Analog In-Memory Computing (AIMC) CUs~\cite{hp_ssc2020, ueyoshi2022diana}. AIMC can be faster and more energy-efficient, but produces approximated results due to the use of extreme quantization bit-width for weights, such as binary or ternary levels. In contrast, digital CUs, while slower and more energy-hungry, handle data with higher numerical precision and thus deliver more accurate results.
Other SoCs~\cite{darkside} are equipped with CUs that can only execute specific DNN layers, such as depthwise convolutions.
Using these CUs implies constraining the DNN to certain architectural patterns, which again can affect the accuracy (usually in exchange for improved efficiency).

In this work, we present \textbf{O}ne-shot \textbf{Di}fferentiable \textbf{M}apping \textbf{O}ptimizer (\textbf{ODiMO}) a novel approach to optimize and map DNN execution onto heterogeneous systems.
In ODiMO, the mapping problem is framed as a \textit{training-time optimization} where both functional (i.e., task performance) and non-functional (i.e., energy efficiency or latency) properties of each CU are taken into account, thus enabling the discovery of accurate yet efficient mappings.
ODiMO currently targets heterogeneous SoCs architectures that present multiple CUs, communicating with each other through a shared on-chip memory (depicted in yellow in Fig.~\ref{fig:intro}). Nonetheless, this work could be easily extended to SoCs with private CUs’ memories by taking into account the cost of broadcasting load and store operations among them.

The main novelties of this work are detailed below:
\begin{itemize}
    \item Our method operates at fine grain, using a gradient-based search technique at training time, to divide each DNN layer into sub-layers, which are then executed in parallel by various CUs, as depicted in Mapping C of Fig.~\ref{fig:intro}. By considering potential accuracy losses due to quantization or layer type selection (e.g., normal convolution vs depthwise convolution), our approach aims to balance the tradeoff between accuracy and energy consumption or latency through the use of analytical and differentiable hardware-aware cost models.
    \item We evaluate the proposed approach on three well-known edge-relevant benchmarks from computer vision, namely CIFAR-10, CIFAR-100, and ImageNet. On these three datasets, we use ODiMO to explore mappings for two state-of-the-art (SotA) open-source heterogeneous SoC designs, i.e., DIANA~\cite{ueyoshi2022diana} and Darkside~\cite{darkside}.
    \item Thanks to our fine-grain mapping discovery procedure we obtain novel solutions that outperform manual heuristic mappings. In particular, ODiMO improves latency by up to $4.9\times$/$8\times$ on DIANA and Darkside respectively, with an accuracy drop $< 2\%$/no-accuracy drop compared to manual heuristic mappings. Similarly, ODiMO improves DIANA's/Darkside's energy efficiency, by up to $1.41\times$/$50.8\times$ with an accuracy drop $< 0.5\%$.
\end{itemize}
The code developed to implement ODiMO and to reproduce the experimental results is open-sourced at \texttt{https://github.com/eml-eda/odimo-journal}. New developments and experiment results will be included at the same link.

This work significantly extends~\cite{odimo_conf} by 
%
%
generalizing the methodology, and proposing novel experiments and comparisons. From the methodological standpoint, we formalize DNN mapping optimizations for heterogenous SoCs with \textit{incompatibilities among CUs} into a general framework, which includes the original case of CUs with incompatible data formats presented in~\cite{odimo_conf}. 
We first show how this special case fits within our framework, and then extend it to a new category of SoCs, with Specialized CUs.
Experimentally, we validate the framework on a new hardware platform, i.e., Darkside, and on two new tasks i.e., CIFAR-100 and ImageNet, demonstrating the discovery of Pareto-optimal mappings in both accuracy-latency and accuracy-energy spaces. We also perform new comparisons between the obtained mappings and SotA alternatives such as Structured Pruning~\cite{pit_journal} and path-based DNAS~\cite{liu2018darts}. Lastly, we detail the search cost of ODiMO in terms of Average Epoch Time and GPU Memory, and perform micro-benchmarking results to validate the developed cost models against real measurements.

The rest of the paper is structured as follows. Sec.~\ref{sec:background} summarizes the required background concepts, while Sec.~\ref{sec:related} covers the most relevant works related to our research. Sec.~\ref{sec:methods} discusses the proposed methodology, which is experimentally validated in Sec.~\ref{sec:results}. Finally, Sec.~\ref{sec:conclusion} concludes the manuscript. 

\section{Background} \label{sec:background}
\subsection{Specialized hardware for edge DNN inference}~\label{sec:soc}
In recent years, specialized architectures for DNN processing at the edge have proliferated, with numerous designs emerging from both industry and academia~\cite{reutherAI2021}.
Many modern SoCs feature multiple specialized CUs capable of executing DNN layers with varying trade-offs in terms of latency, throughput, energy consumption, and accuracy.
For instance, the Jetson AGX Xavier series from NVIDIA is a commercial device including an 8-core ARM CPU, an NVIDIA Volta GPU with 512 CUDA cores, and two NVIDIA Deep Learning Accelerators (NVDLAs). Users can distribute the workload between the GPU, which is faster but more energy-intensive, and the NVDLAs, which are slightly slower but more efficient~\cite{Dagli2022}.
Another commercial alternative is represented by the GAP9~\footnote{\url{https://greenwaves-technologies.com/gap9_processor/}} SoC which includes a cluster of 8 general purpose cores based on the RISC-V ISA along with the Neural Engine 16~\cite{ne16}, a custom CU tailored to accelerate specific convolutional kernels with varying integer precision from 2 to 8 bits.

A similar architecture is used by Darkside~\cite{darkside}, an academic and open-source design. 
Darkside includes a general-purpose control CPU, known as the fabric controller, along with an 8-core cluster of general-purpose RISC-V processors supporting integer arithmetic and a custom CU, the Depthwise Convolution Engine (DWE). In this SoC, the cluster despite being composed of general-purpose cores acts as a de facto flexible CU capable of accelerating generic convolutional workloads. Conversely, the DWE is used to accelerate specifically depthwise convolutions, an operation notoriously characterized by low arithmetic intensity and thus poor performance on general-purpose cores. Therefore, both the DWE and the cluster CUs are not able to execute code stand-alone but function as accelerators, under the orchestration of the fabric controller.
The cluster and the DWE share an L1 memory composed of 32 4-kB SRAM banks capable of serving up to 32 requests in parallel. The memory hierarchy is completed by 256 kB of L2 accessed through a dedicated Direct Memory Access (DMA) co-processor. 
Moreover, Darkside also includes a 16-bit floating-point Tensor Product Engine (TPE) originally proposed to enable on-device learning. 

Another emerging paradigm towards heterogeneity is represented by SoCs including CUs based on Near-Memory Computing (NMC) and Analog In-Memory Computing (AIMC).
In the architecture presented in~\cite{hp_ssc2020}, a control CPU assigns the workload to either a 590k-cell AIMC CU, optimized for 1-bit multiply-and-accumulate (MAC) operations, or a digital NMC CU, which supports variable precision from 1 to 8 bits. Choosing between these two CUs involves a trade-off, as the NMC offers potentially higher accuracy at the cost of increased latency and energy consumption.
Similarly, the DIANA architecture described in~\cite{ueyoshi2022diana} incorporates a single-core RISC-V CPU as control processor, and two distinct DNN-specific CUs. One is a 16$\times$16 grid of digital processing elements that perform MACs at 8-bit precision and include a 64 kB weight memory, while the other is a 500k-cell AIMC CU with ternary weights. Both CUs share a dedicated 256 kB L1 memory with double Read/Write ports, accessed via DMA.

In this work, we consider DIANA and Darkside as target HW platforms as two representatives of the heterogeneous edge SoCs landscape, different from each other.
In particular, DIANA includes two CUs supporting different quantization precision, whereas Darkside includes a unit supporting only a specific layer type (the DWE) along with a more flexible CU (the cluster).
Nonetheless, all techniques discussed in this paper are sufficiently general to be applicable to other platforms with different types and numbers of CUs.
\subsection{Training-time Optimization of DNNs}
Over the years, coupled with the introduction of new specialized SoCs, many DNN optimization strategies have been developed to build lightweight networks appropriate to be deployed on edge devices.
Initially, the design of architectures encompassed the study of efficient operators~\cite{howard2017mobilenets, tan2019efficientnet} along with careful hyperparameter tuning.
Additional methods include quantization~\cite{Jacob2018}, which lowers the precision of the network's weights and activations to minimize memory and computational costs, and pruning~\cite{pit_journal}, which reduces the number of parameters in a network by removing unimportant or redundant ones.
All these optimizations combined create a very large space of design choices. How to efficiently, optimally, and automatically explore this space is an open research problem.

Early approaches towards the automated exploration of DNN optimizations were based upon black-box optimization agents, typically using Reinforcement Learning (RL)~\cite{tan2019mnasnet} or Evolutionary Algorithms (EA)~\cite{evol_nas}, which sampled points in an iterative fashion from the search space. Then, each point was evaluated against functional (e.g., accuracy) and non-functional metrics (e.g., latency, memory footprint) to provide the agent with a reward used to refine subsequent samplings. In this approach, evaluating each candidate solution's accuracy requires a full DNN training, a severe bottleneck hindering its adoption.
Indeed, this strategy scales poorly with the search-space dimension requiring 100s of GPU hours for a single search~\cite{tan2019mnasnet}.
\begin{figure}
  \centering
  \includegraphics[width=\columnwidth]{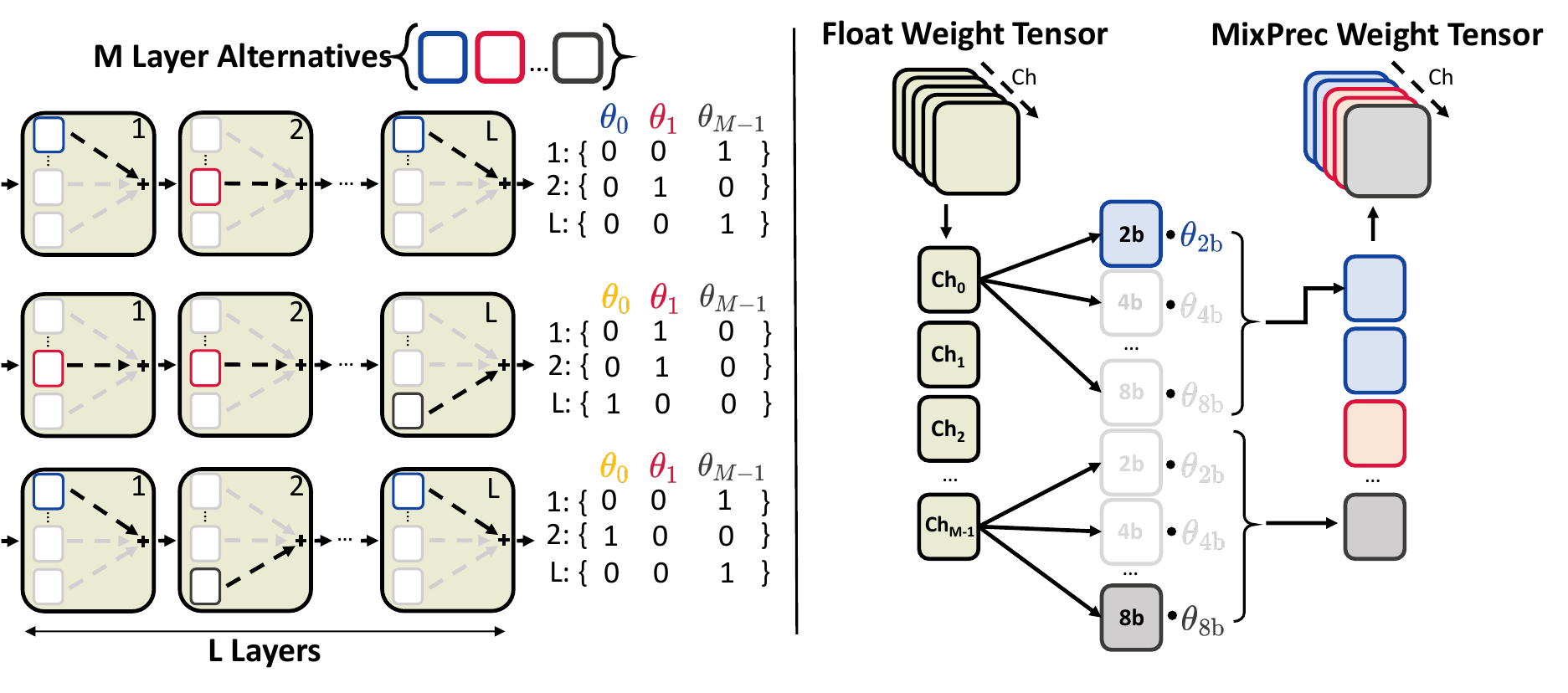}
  \caption{Examples of $\theta$ encoding for Layer Selection (left) and for Mixed-Precision Assignment (right).}
  \label{fig:enc}
\end{figure}

A more recent and time-efficient exploration scheme is based upon parametrizing the search space via a set of trainable parameters $\theta$, and optimizing these parameters using gradient descent as a search strategy~\cite{liu2018darts}.
Each point of the search space is identified by a specific combination of $\theta$ values.
Fig.~\ref{fig:enc} shows two examples of this parametrization. On the left, arrays of $\theta_{i}$ parameters encode the choice between mutually exclusive layer alternatives. On the right, each $\theta_{i}$ is similarly associated with a specific quantization format (e.g., 2-, 4-, 8-bit) for every channel (or filter) of a convolution's weights.
The main benefit of parametrizing the search space in this way is that we can explore it (i.e., optimize the $\theta$ values) in a standard training loop, jointly with the normal network parameters $W$. This can be achieved either through a continuous relaxation of the sampling followed by a discretization at the end of training~\cite{liu2018darts}, or through various forms of differentiable discrete sampling~\cite{duccio}.
Regardless of the sampling scheme, these methods commonly train $W$ and $\theta$ to optimize a loss function in the form:
\begin{equation}
\label{eqn:loss_function}
    \min_{W, \theta} \left[ \mathcal{L}_{\text{task}}(W, \theta) + \lambda \mathcal{C}(\theta) \right]
\end{equation}
where $\mathcal{C}$ is a differentiable cost estimate that takes into account non-functional metrics, $\mathcal{L}$ is the standard task-specific loss function (e.g., cross-entropy loss),  and $\lambda$ is a scalar controlling the trade-off between $\mathcal{C}$ and $\mathcal{L}$. For instance, $\mathcal{C}$ might encode with a differentiable function metrics such as the model size, or the inference latency~\cite{proxylessnas_2018}.

This work tackles the problem of DNNs mapping on heterogeneous CUs as a differentiable optimization, performed at training time, following the scheme detailed above. The problem is framed either as \textit{layer selection} or \textit{mixed-precision assignment}. These two techniques are generally introduced in Sec.~\ref{subsec:layersel_back} and Sec.~\ref{subsec:mixprec_back}, while Sec.~\ref{sec:methods} details how they are applied to the mapping problem studied in this manuscript.
\subsubsection{Layer Selection} \label{subsec:layersel_back}
The first and most straightforward application of the differentiable DNN optimization paradigm is layer selection. In this case, the search space is encoded by building a \textit{supernet}~\cite{liu2018darts} which is a DNN that incorporates multiple alternative paths for each layer, with each path representing a candidate solution. For example, as shown in the left part of Fig.~\ref{fig:enc}, each layer of a reference DNN can be replaced with a module containing various alternatives, such as convolutions with different filter sizes or depthwise vs standard convolutions. During training, the selection of each of these alternatives is associated with trainable parameters $\theta$, optimized according to Eq.~\ref{eqn:loss_function}. After training, a discretization stage selects the best path by combining the alternatives with the highest $\theta$ in each supernet layer. In the literature, this approach is often known as path-based Differentiable Neural Architecture Search (DNAS)~\cite{plinio}. 
\subsubsection{Mixed-Precision Assignment} \label{subsec:mixprec_back}
Mixed-precision quantization refers to an optimization in which different parts of a DNN are quantized to different data formats, possibly providing time, memory, and energy savings with respect to fixed-precision solutions, especially when native hardware support for sub-byte operations is available~\cite{wang2019haq,risso2022channel, ne16, ueyoshi2022diana}. However, the space of possible bit-widths assignments for different parts of the network is huge and exponential with the depth of the DNN. Thus finding an optimal assignment, e.g., to minimize the latency with a certain accuracy constraint is far from trivial.

Some solutions to this problem use sensitivity-based heuristics~\cite{dong2020hawq} or RL~\cite{wang2019haq}. More recently, the differentiable optimization paradigm has been exploited also for this problem~\cite{cai2020rethinking,risso2022channel} where the bit-width assignment is optimized \textit{during training} by solving an optimization problem of the form of Eq.~\ref{eqn:loss_function}.
In this case, as shown in the right part of Fig.~\ref{fig:enc}, the $\theta$ trainable parameters are associated to different versions of the same tensor quantized at different bit widths. Then, an appropriate cost function $\mathcal{C}$ is used to guide the optimization promoting the $\theta$ parameters associated with quantizations that yield a good trade-off between inference cost and accuracy.
At the end of the training, the bit-widths that have been assigned the largest $\theta$ coefficient are selected for each tensor.

\begin{table*}[t]
\centering
\caption{Summary of state-of-the-art Mapping strategies on Heterogeneous Hardware}
\label{tab:related}
\resizebox{\textwidth}{!}{%
\begin{tabular}{lllllll}
\hline
\textbf{Method} & \textbf{Partitioning Engine} & \textbf{Platform} & \textbf{Target Metric} & \textbf{Network Type} & \textbf{Mapping Granularity} & \textbf{Acc. awareness} \\ \hline
Wang \textit{et al.}~\cite{Wang2020} & Fastest First & CPU+GPU+NPU & Latency/Energy & CNN & Entire Network & \ding{55} \\
HDA~\cite{hda} & Heuristic & Multiple NPUs & Latency & CNN & Entire Network & Partial \\ \hline
Vasiliadis \textit{et al.}~\cite{Vasiliadis2022} & RF Scheduler & CPU+GPU & Latency/Energy & CNN & Layer-wise & \ding{55} \\
Tu \textit{et al.}~\cite{Tu2019} & Heuristic & GPU+FPGA & Latency/Energy & CNN &  Layer-wise & \ding{55} \\
AxoNN~\cite{dagli2022axonn} & Linear Programming & GPU+NVDLA & Latency/Energy & CNN & Layer-wise & \ding{55} \\
Jeong \textit{et al.}~\cite{Jeong2022} & Heuristic & GPU+NVDLA & Throughput & CNN & Layer-wise & \ding{55} \\ 
HaX-CoNN~\cite{hax_conn} & SAT Solver & GPU+NVDLA & Latency & CNN & Layer-wise & \ding{55} \\ 
Omniboost~\cite{karatzas2023omniboost} & Monte Carlo Tree Search & CPU+GPU & Throughput & CNN & Layer-wise & \ding{55} \\ 
H3M~\cite{h3m} & Evolutionary & FPGA & Energy-Delay Product & CNN/Transformer & Layer-wise & \ding{55} \\ 
MaGNAS~\cite{magnas} & Evolutionary & GPU+NVDLA & Latency/Energy & GNN & Layer-wise  & \ding{51} \\ \hline
AccPar~\cite{Song2020} & Dynamic Programming & TPU-v2/v3 & \textsc{Training} Latency & CNN & Intra-Layer  & \ding{55} \\ 
Map-and-Conquer~\cite{mapconq} & Evolutionary & GPU+NVDLA & Latency/Energy & CNN/Transformer & Intra-Layer & Partial \\ \hline
\textbf{Ours} & \textbf{Gradient-based} & \textbf{DIANA, Darkside} & \textbf{Latency/Energy} & \textbf{CNN} & \textbf{Intra-Layer} & \ding{51} \\ \hline

\end{tabular}
}
\end{table*}

\section{Related Works} \label{sec:related}
How to efficiently map computationally expensive workloads onto the CUs available on heterogeneous systems represents a compelling problem. Initial studies concentrated on general-purpose workloads, such as OpenCL programs~\cite{Konrad2018}. However, in recent years, there has been a growing interest in the more specific domain of DNN inference.
In~\cite{Wang2020}, a mobile SoC composed of CPU, GPU, and NPU is considered. At any time the fastest available CU is selected to map the entire DNN, parallelizing multiple inference requests. 
Similarly, HDA~\cite{hda} considers the case of an edge inference server equipped with custom NPUs with 8, 7, and 6-bit quantization. In this case, a scheduling approach is proposed to map entirely and concurrently multiple DNN inference requests to minimize latency. Namely, when a new request arrives, the fastest available NPU that satisfies a certain accuracy threshold is selected. This accuracy threshold is defined task-wise and takes into account the resiliency of the considered DNNs to quantization (profiled offline).
\cite{Vasiliadis2022} explores DNN mapping on CPU and GPU at the granularity of single layers. At runtime, layers are mapped to CUs to minimize the overall inference energy or latency using a Random Forest (RF) predictor, that uses the layer's hyper-parameters as input features.
\cite{Tu2019} profiles DNNs execution over a multi-CU system including a GPU (NVIDIA Jetson TX2) and an FPGA (Xilinx Artix7), coming up with a heuristic mapping strategy consisting of offloading the whole network to the GPU, except for Fully-Connected (FC) layers which are executed on the FPGA.
AxoNN~\cite{Dagli2022} considers partitioning DNNs at the layer level on an NVIDIA Jetson AGX Xavier. Using a linear programming approach, the energy versus latency trade-offs offered by offloading parts of a DNN to the GPU or the NVDLAs are explored. \cite{Jeong2022} proposes an alternative mapping scheme for the same platform, where the focus is improving throughput by exploiting data parallelism and pipelining among GPU and NVDLAs.
HaX-CoNN~\cite{hax_conn} uses a SAT solver to optimize the latency of the interleaved execution of multiple DNNs, mapped at the layer granularity over the CUs of an NVIDIA Jetson TX2. Similarly, Omniboost~\cite{karatzas2023omniboost} proposes an layer-wise mapping scheme based on Monte Carlo Tree Search for serving multi-DNN workloads to minimize throughput on a platform including 2 CPUs and a GPU.
H3M~\cite{h3m} targets datacenter FPGAs with an evolutionary-based strategy to map onto the HW multiple DNNs with layer granularity to minimize energy-delay product.
MaGNAS~\cite{magnas} explores the per-layer mapping of Graph Neural Networks (GNNs) over GPU and NVDLAs. Concurrently the architecture of the GNN is also optimized. Both mapping and architecture are optimized using an evolutionary algorithm to maximize accuracy on the considered task while reducing the energy or latency.

All these previous works consider \textit{coarse mappings} where the atomic element that is assigned to a certain CU is an entire layer.
Conversely, AccPar~\cite{Song2020}, explores finer-granularity intra-layer partitions. Taking into account compute performance and communication overheads, dynamic programming is used to optimize DNN training latency over a cluster of multiple Google TPU-v2/v3. The considered partitioning axes are: over batches (i.e., data parallelism), input channels, or output channels.
Targeting inference, instead, Map-and-Conquer~\cite{mapconq} explores intra-layer mappings considering again a Jetson AGX Xavier as the target platform. An evolutionary algorithm is used to explore different partitioning schemes to find the optimal energy-delay product. The intra-layer mapping is achieved at the channel level in Convolutional Neural Networks (CNN) and the head level in Vision Transformers (ViT) by neglecting the data dependencies between contiguous groups of weights. This operation is performed post-training considering the optimization objective defined in~\cite{Molchanov_2019_CVPR}, originally proposed for pruning, thus the impact on accuracy is only partially taken into account. Moreover, as in MaGNAS~\cite{magnas}, the accuracy effect is not related to specific characteristics of the CUs, but only to the architecture of the network.

Conversely, our method considers \textit{fine-grained intra-layer mappings} aimed at optimizing any cost metric such as energy or latency while being \textit{completely accuracy-aware}.

We report in Table~\ref{tab:related} a summary of the state-of-the-art mapping schemes on heterogeneous platforms.
Importantly, to our knowledge, ours is the first approach in the literature to consider a gradient-descent-based method to optimize the mapping/partitioning of DNN computations onto multiple hardware CUs.

\section{Proposed Method} \label{sec:methods}
The majority of the related works discussed in Sec.~\ref{sec:related} are accuracy-unaware, and only explore the trade-off between latency and throughput or latency and energy. Therefore, they do not consider heterogeneous platforms where the execution on different CUs may affect the final task accuracy.
This is the case of new edge-oriented platforms such as~\cite{hp_ssc2020,ueyoshi2022diana, darkside} where due to different data representations or limitations in the accelerable operations (e.g., depthwise vs standard convolution) accuracy can be heavily impacted by mapping choices.
For these reasons, existing methods cannot be applied to such new hardware straightforwardly.
HDA~\cite{hda} which is the sole strategy aware of the possible accuracy impact of different CUs, addresses the different problem of serving multiple inference requests concurrently. In contrast, we concentrate on optimizing a single, unbatched inference, which is the common case in extreme-edge devices. Indeed, these devices usually process newly available inputs immediately, often in real-time.

To the best of our knowledge, ODiMO represents the first DNN mapping tool tailored for SoCs where CU heterogeneity directly impacts accuracy. Namely, ODiMO explores, during training, the accuracy vs latency/energy trade-off determining an optimal DNN partitioning through hyper-parameters optimization (e.g., layer type or quantization precision).

In practice, our method stems from the observation that, for systems including heterogeneous CUs with the incompatibilities described above, \textbf{optimizing a DNN's architecture (e.g. layer type or quantization format), and determining its mapping onto the various CUs, reduce to the same problem}. In fact, assigning a certain operation or data format to part of the DNN, automatically means that such part will be executed on the specific CU(s) that supports it. Therefore, we can adapt existing gradient-based DNN optimization techniques to address the heterogeneous mapping problem.

In contrast to prior strategies that limit mappings to coarse, layer-wise assignments (entire layers on one CU), ODiMO supports fine-grained intra-layer partitioning. This granularity improves utilization across all CUs.

The rest of this section is organized as follows. Sec.~\ref{subsec:strategy} formalizes the network optimization/mapping problem and the proposed ODiMO strategy. Sec.~\ref{subsec:data_formats} shows how to use ODiMO to map DNNs onto CUs characterized by incompatible quantization formats, while Sec.~\ref{subsec:hw_units} presents the case of mapping onto SoCs with specialized HW units.
\subsection{Mapping optimization strategy}\label{subsec:strategy}
\begin{figure}[t]
  \centering
  \includegraphics[width=\columnwidth]{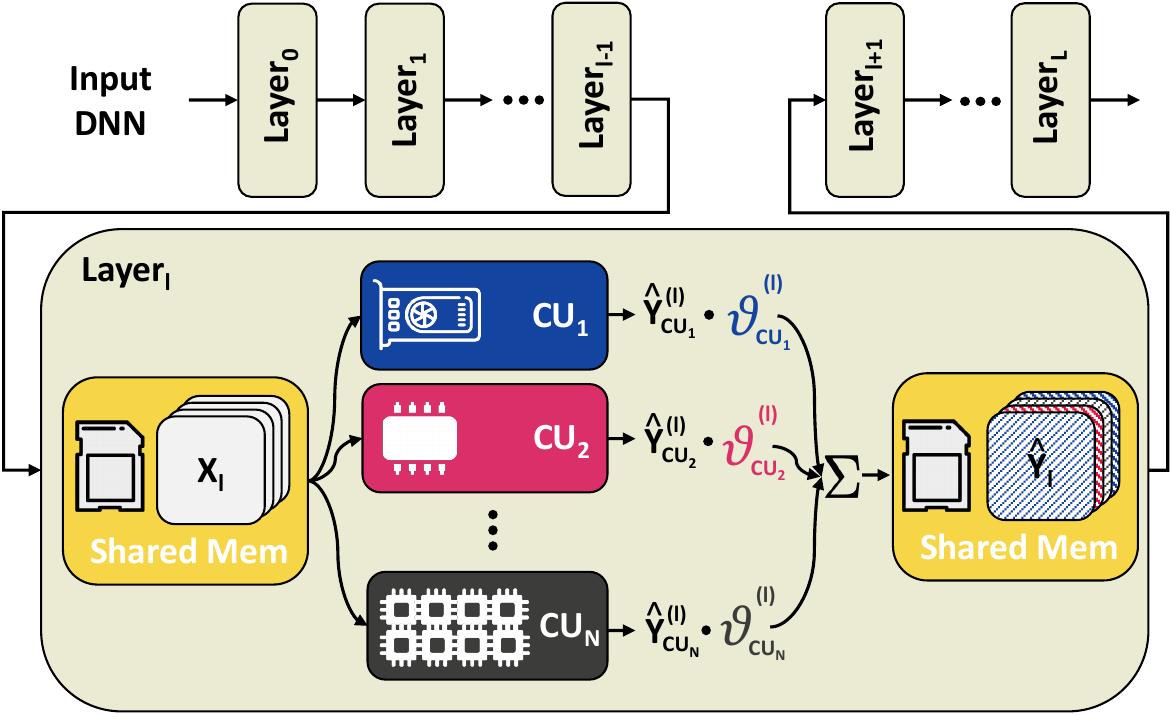}
  \vspace{-0.2cm}
  \caption{General ODiMO mapping strategy for a layer.}
  \label{fig:map}
\end{figure}
Given a SoC with $N$ different Computing Units (CUs) and a DNN, ODiMO explores how to partition each Convolutional (Conv) or FC layer, at the level of \textit{individual output channels/neurons}, among the available CUs. In the following discussion, we will always use the term output channels, without loss of generality. Fig.~\ref{fig:map} shows an example of the mapping operated by ODiMO for a Conv layer: all CUs take the entire layer's input, and produce a subset of the output activations' channels.
As anticipated in Sec.~\ref{sec:intro}, we consider heterogeneous systems where all the CUs can access a shared memory region (represented in yellow in Fig.~\ref{fig:map}) for loading/storing the layer input/partial output. 
By partitioning over output channels, each CU processes a different set of weights (which can be stored privately), and stores its outputs to different memory locations. Each input activation, instead, is loaded $N$ times, once per CU. Note that this data transfer redundancy is taken into account by our optimization's cost models (see below), and remains beneficial despite its overheads, as it enables parallel usage of multiple CUs.

We assume that the shared activations memory is multi-ported and multi-banked and provides enough bandwidth to allow concurrent usage by all CUs, with accesses to the same location resolved by an arbitration mechanism. These requirements are met by many different real-world designs such as~\cite{Song2020, ueyoshi2022diana, hp_ssc2020, darkside}. Nonetheless, as said before, this work could be easily extended to CUs with private memories by considering the overhead of broadcasting the required input data to them. We will address this scenario in our future works.

Given this fine-grain intra-layer mapping, the size of the search space explodes: e.g., for just $N=2$ CUs and a ResNet18 CNN, there are about $10^{39}$ possible ways to assign each channel of each layer to one of the two units.
To efficiently explore a search space of such size, ODiMO adopts a differentiable optimization strategy, where the possible mappings are parametrized through trainable parameters $\theta$. 

In this way, by solving an optimization problem similar to the one of Eq.~\ref{eqn:loss_function} we can efficiently explore mappings \textit{while training} the DNN, balancing cost and task-accuracy in \textit{one-shot}.
In particular, ODiMO is inspired by recent work on differentiable fine-grained mixed-precision quantization~\cite{risso2022channel} and differentiable layer-selection~\cite{liu2018darts}.

As shown in Fig.~\ref{fig:map}, first, ODiMO identifies the layers $l$ that can be mapped onto the available CUs. 
Then, it simulates the effect of offloading the computation of \textit{every entire layer's output $\hat{Y}_{i}^{(l)}$ to \textit{every} CU}. The result is a set of $N$ possible outputs $\left\{ \hat{Y}_{CU_{j}}^{(l)} \right\}_{j}^{N}$.
Finally, the \textit{effective} output feature map $ \hat{Y}^{(l)} $ is built by forming each of the $C_{out}^{(l)}$ output channels as a linear combination of the output produced by each CU, weighted by the set of trainable parameters $\theta = \left\{ \theta^{(l)}_{CU_{j}, \: c} \right\}_{j, \: c}^{N, \: C_{out}^{(l)}} $.
In this way, ODiMO builds each layer's output as a mixture of what would be produced by all available CUs, given their computing scheme. 
Mathematically, we compute each output channel $c$ of each layer $l$ as:
\begin{equation} \label{eq:weighted_output}
    \hat{Y}^{(l)}_{c} = \sum_{j}^{N} \theta^{(l)}_{c, \text{CU}_j} \hat{Y}^{(l)}_{c, \text{CU}_j}
\end{equation}
For each channel $c$ in layer $l$, ODiMO sets $\theta^{(l)}_{c,\text{CU}_{j}} = 1$ when assigning that channel to CU $j$ provides improved accuracy and energy/latency trade-off . Conversely, all other assignments for that channel ($\theta^{(l)}_{c, \text{CU}_{k}}$ where $k \ne j$) are set to zero.
During the optimization, the values of $\theta$ can be either sampled discretely~\cite{duccio} or relaxed to assume continuous values in  [0,1], e.g., by applying a $\mathrm{softmax}$ operator to a vector of free trainable parameters $\bar{\theta}$~\cite{liu2018darts}. At the end of the training, the CU whose $\theta_{CU_{j, c}}$ is associated with the largest sampling probability (or value, in the case of a continuous relaxation) is selected as the final offloading target for the $c$-th channel.
After the final assignment of channels to CUs is determined, the layer is reorganized into N parallel sub-layers, each comprising the subset of channels that have been assigned to the $j$-th CU. These $N$ sub-layers can be executed in parallel, and their outputs will be concatenated in the shared memory to be used as inputs for the next layers. More details on this conversion process are provided in Sec.~\ref{subsec:data_formats}.

The assignment of channels is optimized by training the DNN, modified as detailed above with the additional $\theta$ parameters, to minimize the loss function of Eq.~\ref{eqn:loss_function}. Namely, the $\theta$ and the normal weights $W$ are trained jointly.
In the loss function, the cost term $\mathcal{C}(\theta)$ can change depending on the optimization's non-functional goal. The formulation can be extended to any cost metric that can be expressed as a differentiable function of $\theta$.
For example, when considering latency, the objective of ODiMO is to minimize:
\begin{equation}\label{eq:reg_loss}
    \mathcal{C} = \sum_{l} M^{(l)}\mathrm{,\ }M^{(l)} = \max(LAT^{(l)}_{1}(\theta),...,LAT^{(l)}_{n}(\theta))
    \vspace{-0.2cm}
\end{equation}
where each $LAT^{(l)}_{i}(\theta)$ is a differentiable model, parametrized by $\theta$, of the $l$-th layer's execution latency on the $i$-th CU, as a function of the channels assigned to it. $M^{(l)}$ is the latency of the entire layer, computed with a $\mathrm{max}()$ operation since we assume that all CUs run in parallel. We consider this scenario because minimizing the idleness represents the optimal choice for both time and energy reduction. In practice, since we need a fully-differentiable loss term, we substitute the $\mathrm{max}$ operation of Eq.~\ref{eq:reg_loss} with its smooth differentiable approximation, computed as the sum of the different terms weighted by the corresponding $\mathrm{softmax}$-ed parameters. 
For energy reduction, instead, we use the following model:
\begin{equation}\label{eq:reg_loss_en}
    \mathcal{C}_{en} = \sum_l (\sum_i P_{act, \: i}\cdot LAT^{(l)}_{i}(\theta)) + P_{idle}\cdot M^{(l)} 
    \vspace{-0.2cm}
\end{equation}
This model accounts for both active and idle power consumption across all DNN layers. For each layer $l$, we compute: (i) the active energy consumption, i.e., $\sum_i P_{act, \: i}\cdot LAT^{(l)}_{i}(\theta)$, where, $P_{act, \: i}$ represents the average active power consumed by the $i$-th CU during computation, beyond its idle state. This is multiplied by the execution time $LAT^{(l)}_{i}(\theta)$ of the channels assigned to that CU. (ii) the baseline energy consumption: $P_{idle}\cdot M^{(l)}$. This term captures the platform's idle power ($P_{idle}$) over the total latency of the layer ($M^{(l)}$). The total energy is then accumulated across all layers through the outermost summation.
The power values involved in Eq.~\ref{eq:reg_loss_en} can be measured on the HW with simple profiling runs or taken from a datasheet, depending on the required modelling accuracy.

Regardless of the specific hardware platform considered, ODiMO entails three training phases to generate an optimized mapping. In the first phase (\textit{Warmup}), the mapping parameters $\theta$ are kept frozen, and the network is trained to minimize the task loss $\mathcal{L}$ only (without considering cost) through the $W$ parameters.
The rationale is that starting the mapping optimization from a well-trained network ensures a more accurate task-performance evaluation and ranking of solutions. In contrast, directly optimizing an untrained network would bias the search towards lower-cost (e.g., lower-latency) solutions, due to the fact that the task performance of all available alternatives is ``equally bad'' at the beginning of the training.
Then, we move to the \textit{Search} phase where cost $\mathcal{C}$ and task loss $\mathcal{L}$ are jointly optimized as in Eq.~\ref{eqn:loss_function}, and both the $\theta$ and the standard weights $W$ are trained, in order to find a well-performing assignment of layer portions to CUs.
Finally, during the \textit{Final Training} phase, the assignment of channels to CUs is frozen based on the values of the $\theta$ parameters at the end of the search, and, similarly to warm up, the $W$ weights are trained for some more epochs to optimize $\mathcal{L}$ alone. This phase allows the model to recover possible accuracy drops due to the final discretization of the mapping.

During the \textit{Search} phase, the scalar value $\lambda$ of Eq.~\ref{eqn:loss_function} is used as a knob to control the trade-off between the task loss $\mathcal{L}$ and the cost $\mathcal{C}$. Namely, large values of $\lambda$ will prioritize solutions with a low non-functional cost $\mathcal{C}$, while smaller values will give more importance to the task loss $\mathcal{L}$. Repeating the optimization varying $\lambda$, ODiMO can generate a complete Pareto front in the accuracy versus cost space.
\subsection{SoC with Incompatible Data Formats}~\label{subsec:data_formats}
In this section, we describe a first practical instance of the ODiMO scheme for SoCs with multiple CUs with incompatible quantization. An example of this kind of SoC is represented by DIANA~\cite{ueyoshi2022diana} which includes a Digital and an Analog CU with respectively 8bit and ternary precision for the weights.
Other examples include the custom variable precision NPUs considered in HDA~\cite{hda} or the SoC presented in~\cite{hp_ssc2020}, where an NMC CU is coupled with an AIMC one.

In this case, the problem of selecting which CU will execute each layer's channel can be transformed to a \textit{precision assignment problem}, similar to the one tackled in~\cite{risso2022channel}. Differently from standard precision assignment, however, in this case, the selected precision does not only influence the model accuracy, but also the inference energy/latency costs, by implicitly limiting the mapping options for that channel to the CU(s) supporting the selected precision.
For example, in the case of DIANA, assigning ternary precision to the $c$-th channel automatically means that it will be executed on the AIMC CU, with a certain associated cost. Similarly, assigning a channel to 8-bit precision implicitly corresponds to mapping its execution onto the digital CU, with a different impact on latency/energy, and accuracy.
Therefore, with reference to the general ODiMO formulation of Sec.~\ref{subsec:strategy} (Eq.~\ref{eq:weighted_output}), in this case the different CU outputs $\hat{Y}^{(l)}_{c, \text{CU}_j}$ represent the results of \textit{different versions of the same layer, with weights quantized at different precision.}
Then, using the 3-phase training scheme discussed above, ODiMO optimizes the assignment of each channel to a given quantization bitwidth, and therefore, to the corresponding CU. 

\begin{figure}[t]
  \centering
  \includegraphics[width=\columnwidth]{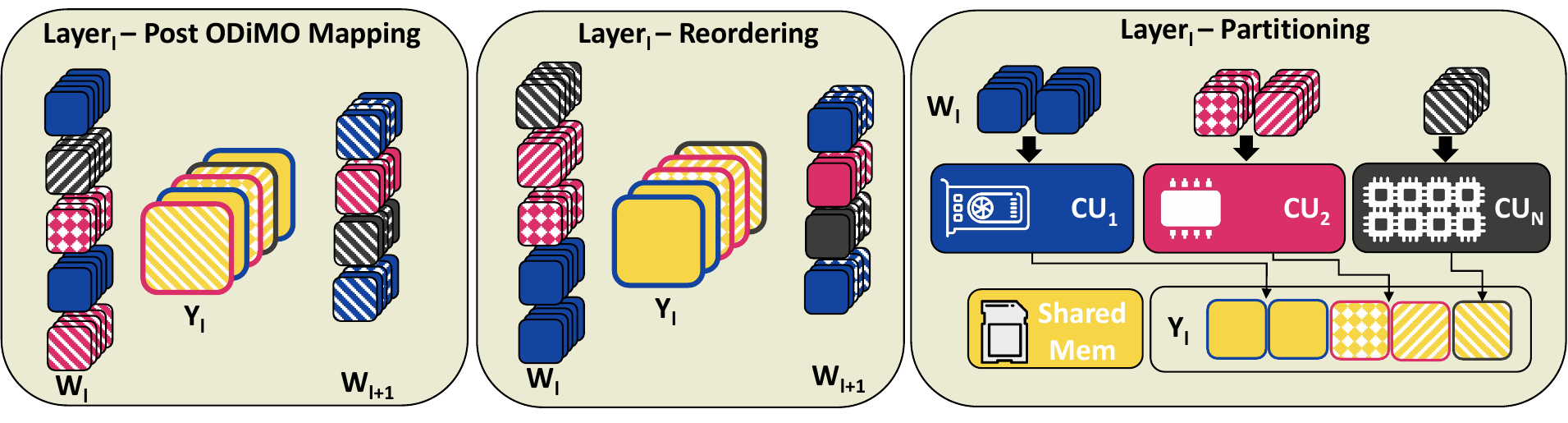}
  \vspace{-0.2cm}
  \caption{Final layer re-organization pass to support partitioning.}
  \label{fig:ch-map}
  \vspace{-0.6cm}
\end{figure}

However, one important caveat is that, in the optimization output, the channels assigned to a given CU for each layer are not ordered. For instance, for a layer with $C_{out}^{(l)}$=8 and two CUs,  channels 0, 3, and 5 could be assigned to $\text{CU}_0$, and channels 1, 2, 4, 6, and 7 to $\text{CU}_1$.  Deploying the layer as is could lead to an inefficient implementation, as different CU outputs will be interleaved in the shared memory. A more efficient alternative is represented by grouping all channels associated to the same CU in contiguous output locations. Therefore, Fig.~\ref{fig:ch-map} shows a layer transformation pass applied to the DNN after the optimization, and before the deployment on the target SoC.
We show it for a Conv layer, as an example, representing activation channels side-by-side with squares, and weight filters as ``cubes". The colors used for the shapes' outlines encode the assignment to a certain CU.
Color patterns are added to some filters/output slices to clarify the process.
The starting point is an ODiMO output, depicted in the top-left part of the figure.
We then group the channels in $Y^{(l)}$ and the corresponding filters in $W^{(l)}$ that will be dispatched to the same CU. Moreover, the weights of the next layer $W^{(l+1)}$  are also reordered across the \textit{input} channels dimensions to preserve the network functionality. This process is depicted in the middle part of Fig.~\ref{fig:ch-map}.
Finally, as shown in the right part of Fig.~\ref{fig:ch-map}, $N$ independent sub-layers, executable in parallel, are obtained by splitting the original one. This step enables the deployment of the obtained mapping onto the $N$ available CUs, without requiring any data-marshaling overhead to aggregate their outputs.

Note that, in the previous part of this section, we presented the case of CUs with incompatible weight quantization formats as a specific case of the general ODiMO mathematical framework  (Eq.~\ref{eq:weighted_output}).  While this formulation is perfectly valid, our actual implementation is different, purely for training efficiency reasons. 
Namely, rather than combining the \textit{output activations} produced by layers with differently quantized weights, we exploit the linearity of Conv/FC operations to \textit{directly combine the weights}. Namely, we exploit the following factorization, equivalent to Eq.~\ref{eq:weighted_output}:
\begin{equation} \label{eq:weighted_output_expanded}
    \hat{Y}^{(l)}_{c} = (\sum_{j}^{N} \theta^{(l)}_{c, \text{CU}_j} W^{(l)}_{c, \text{CU}_j}) * X^{(l)}
\end{equation}
where $W^{(l)}_{c, \text{CU}_j}$ is the c-th weights kernel of the l-th layer, quantized at the precision supported by the j-th CU.
The part inside parentheses represents an \textit{effective weights filter}.
The advantage of this alternative formulation is that it does not require computing multiple separate convolutions for each layer. Rather, it just requires constructing the effective weights, through simpler element-wise operations. Therefore, it reduces the execution time of ODiMO. 
\subsection{SoC with Specialized HW Units}~\label{subsec:hw_units}
This section presents how to tailor ODiMO to CUs that are compatible from the data format standpoint, but support the execution of different types of layers. This is the case of SoCs such as Darkside~\cite{darkside} where a general-purpose multi-core cluster of RISC-V processors is coupled with a CU specialized to efficiently execute depthwise convolutions, the DepthWise Engine (DWE). Another example that fits this scheme is Kraken~\cite{kraken}, a RISC-V heterogenous SoC that includes a multicore cluster, and a Sparse Neural Engine to execute event-based spiking DNNs.
Usually, layers such as normal and DW Convolutions are alternated sequentially, leading to a situation in which only one CU is used at any time. Exploiting the ODiMO optimization framework, we can instead execute multiple sub-layers in parallel, e.g., one normal Conv and one DW, each with fewer output channels. The fraction of channels produced by each sub-layer can be optimized with ODiMO, considering both the relative speeds (or energy consumptions) of the CUs, and the impact of each type of layer on accuracy. As for the case of quantization, executing standard and DW Convs in parallel exposes an interesting trade-off, i.e., a DW Conv will encompass much fewer OPs with respect to a standard one but it will probably impact accuracy more.

More formally, we use once again the scheme of Eq.~\ref{eq:weighted_output}, where this time, $Y^{(l)}_{c, \text{CU}_j}$ and $Y^{(l)}_{c, \text{CU}_k}$ with $k \ne j$ represent two (possibly different) tensor operations, as supported by each CU.
While this scheme can in general accommodate different types of layers (e.g., convolutions with different filter sizes), we test it in practice for the combination of normal and DW convolutions.

Also in this case, the raw ODiMO output has to be slightly altered to produce practically usable DNNs. Namely, we need once again to assign consecutive channels to the same CU. In general, if we map the first $n_c$ channels to the DW conv, then the trailing $C_{out} - n_c$ channels should be executed as a normal convolution.
As for the quantization case, this constraint is needed to avoid costly data marshaling operations during the actual execution of the workload on the SoC. However, differently from what is discussed in Sec.~\ref{subsec:data_formats}, such ``grouping'' cannot be achieved through a post-optimization step. In fact, due to the constraints of DW layers (each output channel being a function of just one input channel), the presence of two or more consecutive layers of this kind, each enforcing a different reordering, would make a transformation like the one of Fig.~\ref{fig:ch-map} impossible.
In this case, to ensure that all channels assigned to the same CU remain ``grouped", we constraint the optimization process. Instead of applying the softmax independently to each element in the $\theta$ array, we first combine them as follows:
\begin{equation}
    \theta^{(l)}_{i} = \sum_{j = 1}^{N - i} \hat{\theta}^{(l)}_{N - j}
\end{equation}
This formulation enforces that if $i > j$, then $ \theta^{(l)}_{i} \le \theta^{(l)}_{j}$. This, in turn ensures that the channels mapped to the same CU are always contiguous.
Despite this additional constraint in the optimization, as shown in Sec.~\ref{sec:results}, ODiMO is still able to discover Pareto-optimal mappings that exploit the HW better than manual mappings.

\section{Experimental Results} \label{sec:results}
\begin{figure*}[t]
  \centering
  \includegraphics[width=\textwidth]{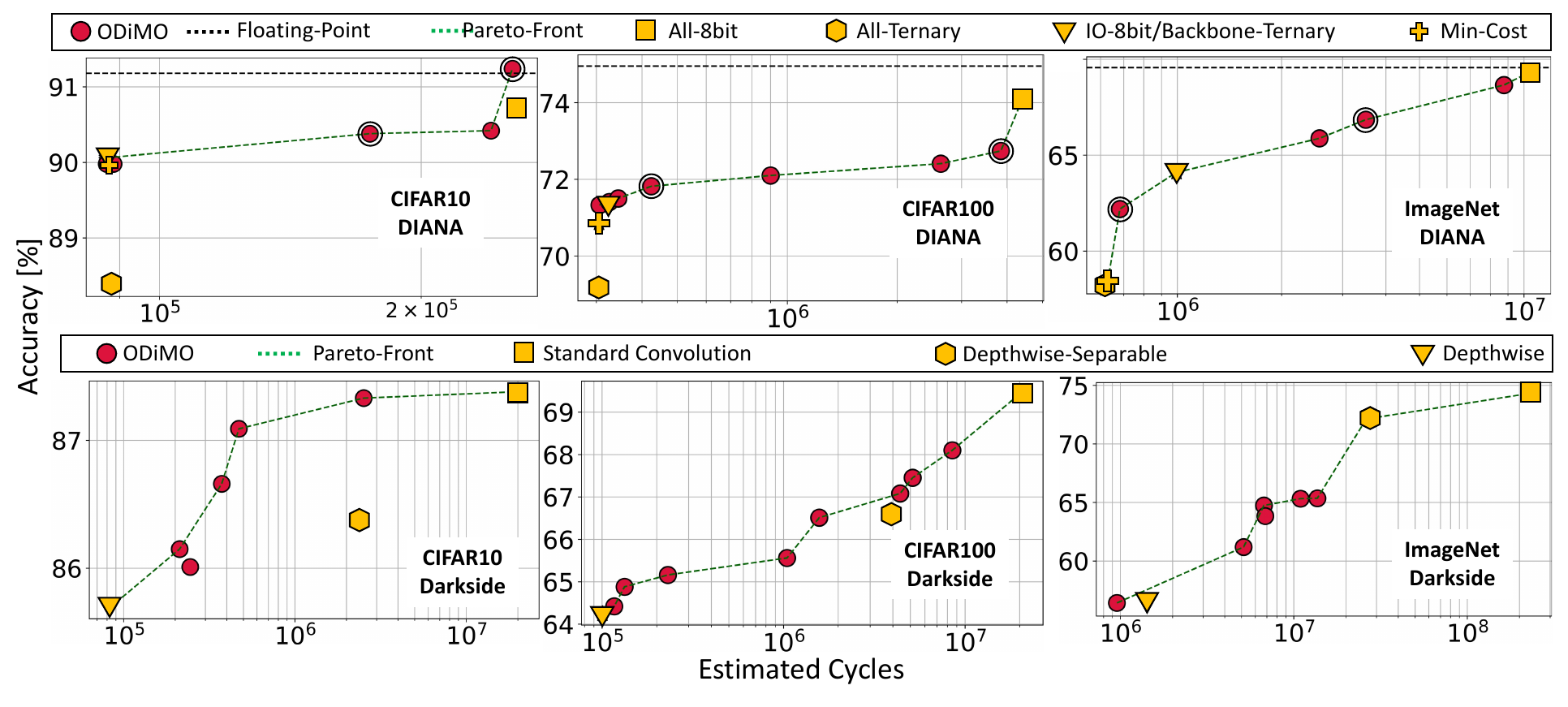}
  \vspace{-.7cm}
  \caption{Mappings obtained using ODiMO with latency as optimization target.}
  \label{fig:fronts}
\end{figure*}
\subsection{Setup}~\label{subsec:setup}
We benchmark ODiMO on three edge-relevant image classification datasets:
i) CIFAR-10~\cite{cifar}; 
ii) CIFAR-100~\cite{cifar}; 
iii) ImageNet-1k~\cite{imagenet}.
We target two different platforms, namely Diana~\cite{ueyoshi2022diana} and Darkside~\cite{darkside} as examples of SoCs including CUs with incompatible data formats (case of Sec.~\ref{subsec:data_formats}) and supporting different layer types (Sec.~\ref{subsec:hw_units}), respectively.

One key element of our methodology that changes for different hardware targets is the expression of the cost term $\mathcal{C}$ in the optimization objective (Eq.~\ref{eqn:loss_function}).
Moreover, the details of the Warmup-Search-Final Training protocol described in Sec.~\ref{subsec:strategy}, also need to be slightly modified depending on the target, especially for the Warmup phase. For details on both these platform-specific customizations, we refer the readers to our \href{https://github.com/eml-eda/odimo-journal}{open-source repository}.

For each dataset, we use as blueprint for our optimization DNNs well-suited to be mapped either on DIANA or on Darkside.
For Darkside and for all three benchmarks we considered MobileNetV1~\cite{howard2017mobilenets} both as a strong baseline with DepthWise-separable convolutions that can benefit from the DWE, and as the starting point to build the supernet used within ODiMO. In particular, starting from the original MobileNet architecture, we let ODiMO optimize a supernet that includes two alternatives (normal Conv and DW Conv) for each layer that has $C_{out} = C_{in}$. Therefore, the mapping solutions discovered by ODiMO will be novel architectures where SW and standard convolutions can be executed in parallel, which is in contrast to the sequential execution scheme of MobileNetV1.
Conversely, for the DIANA platform, which is not optimized to execute depthwise layers (depthwise convolutions can only be executed on the digital CU, and with lower efficiency w.r.t. standard conv), we adopted networks of the ResNet~\cite{resnet} family. 
Nonetheless, the method can be applied to any DNN made of combinations of convolutional and/or linear layers.
In particular, we employed a ResNet20 for CIFAR-10 and a ResNet18 for both CIFAR-100 and ImageNet.
ODiMO is implemented in Python 3.10 and PyTorch v2.3, extending the PLiNIO~\cite{plinio} DNN optimization library.

We compare ODiMO with several baseline mapping alternatives depending on the considered HW. On DIANA, we consider the following baselines: i) \textit{All-8bit} and \textit{All-Ternary}, which are simple mappings utilizing exclusively the digital and AIMC CUs, respectively; ii) \textit{IO-8bit/Backbone Ternary}, a heuristic method proposed in~\cite{ueyoshi2022diana} that assigns the first and last layers to the 8-bit CU and the intermediate layers to the AIMC CU, based on the general guideline that aggressively quantizing the layers near the input and output can significantly impair accuracy; iii) \textit{Min-Cost}, an optimized deterministic mapping that employs the same channel-wise partitioning as ODiMO, aiming solely at minimizing cost, without considering accuracy. Specifically, it statically assigns the channels of each layer to the AIMC and digital CUs before training, intending to minimize Eq.\ref{eq:reg_loss} or Eq.\ref{eq:reg_loss_en}. If multiple solutions yield equivalent costs, digital channels are maximized as this is expected to enhance accuracy.
Therefore, this baseline corresponds to the mapping that achieves the best load-balancing between the available CUs.
Moreover, in Sec.~\ref{subsec:abl_stud}, we compare the solutions obtained using ODiMO with the PIT~\cite{pit_journal} structured channel pruning approach on the CIFAR-10 benchmark.

On Darkside, we consider three baselines. The first is the standard MobileNet with Depthwise-Separable convolutions, i.e., alternations of DW and pointwise convolutions (standard Conv with 1x1 filters).
Then, we consider the case of layers entirely mapped either on the cluster or on the DWE. When the entirety of the network is mapped on the cluster we always execute $3 \times 3$ standard convolution in place of the original Depthwise-Separable, while in the case of the DWE, a $3 \times 3$ DW convolution is used.
Additionally, in Sec.~\ref{subsec:abl_stud}, we show how ODiMO applies to MobileNets initialized with different width multipliers, and we compare the solutions obtained with our method on CIFAR-10 with a standard path-based DNAS approach where entire layers are mapped to one of the available CUs.

\begin{table}[h]
\caption{Average Epoch Time and GPU Memory Overheads of ODiMO Search with respect to the most demanding Baseline on DIANA and Darkside}
\label{tab:overheads}
\resizebox{\columnwidth}{!}{
\begin{tabular}{c|c|c|c}
 & \textbf{Platform} & \textbf{Avg Epoch Time} & \textbf{Memory Overhead} \\ \hline
\multirow{2}{*}{\textbf{CIFAR-10}} & DIANA & 2.3$\times$ & 1.07$\times$ \\ \cline{2-4} 
 & Darkside & 1.99$\times$ & 1.03$\times$ \\ \hline
\multirow{2}{*}{\textbf{CIFAR-100}} & DIANA & 2.48$\times$ & 1.09$\times$ \\ \cline{2-4} 
 & Darkside & 2.12$\times$ & 1.31$\times$ \\ \hline
\multirow{2}{*}{\textbf{ImageNet}} & DIANA & 1.55$\times$ & 1.07$\times$ \\ \cline{2-4} 
 & Darkside & 1.42$\times$ & 1.24$\times$
\end{tabular}
}

\end{table}
Table~\ref{tab:overheads} reports the ODiMO execution overheads for the three considered tasks in terms of average time per epoch and GPU memory occupation, when dealing with the search spaces associated to the two considered hardware platforms. All the overheads are computed with respect to the most demanding baseline for each task and SoC, i.e., the ``All-8bit" baseline for DIANA and the ``Standard Convolution" baseline for Darkside. The time overhead refers to the search phase, and is averaged over 10 epochs. As shown in the table, it spans from $1.42\times$ to $2.48\times$ with an average of $1.93\times$. Notice that this result is in line with the fact that for both the DIANA and Darkside search spaces, ODiMO ``simulates'' the execution of each layer onto two different CUs. Therefore, it needs to perform two forward and backward passes for each layer, which leads to an average 2$\times$ time increase. The memory overhead spans from $1.03\times$ to $1.31\times$ ($1.14\times$ on average) with a peak memory requirement of 8114~MB in the case of the ImageNet task.
\subsection{Training Hyper-Parameters}
For each of the three tasks, we use different training hyper-parameters. We report the main choices here for reproducibility, referring the reader to our \href{https://github.com/eml-eda/odimo-journal}{open-source code} for the remaining details.

We set the training epochs for the warmup, search, and fine-tuning phases to 500, 200, and 130 for the CIFAR-10, CIFAR-100, and ImageNet benchmarks, respectively. 
We apply early-stopping using as control metric the validation accuracy with patience equal respectively to 50, 100, and 30 epochs for CIFAR-10, CIFAR-100, and ImageNet.
For each benchmark, the standard cross-entropy loss is used as task loss $\mathcal{L}$.

On DIANA, we use the SGD optimizer for the weights $W$, with a learning rate of 1e-2, a momentum of 0.9, and a weight-decay of 1e-4. Instead, for the $\theta$ parameters, we use the Adam optimizer with a learning rate of 1e-3.
On Darkside, we use the Adam optimizer for both kinds of optimization parameters with a learning rate of 1e-3 for CIFAR-10 and CIFAR-100 and of 1e-4 for ImageNet.
\subsection{Search-Space Exploration}\label{sec:exploration}
\subsubsection{Latency Optimization}
Fig.~\ref{fig:fronts} presents the results obtained with ODiMO on the three benchmarks in the accuracy versus estimated latency for DIANA (top row) and Darkside (bottom row), with latency estimated using the cost models described in our \href{https://github.com/eml-eda/odimo-journal}{open-source repository}. All the reported accuracies are on the test set with Pareto points selected on the validation set.
Each ODiMO point is obtained varying the regularization strength $\lambda$ of Eq.~\ref{eqn:loss_function}. We also report the baselines discussed in Sec.~\ref{subsec:setup} in green and the floating point DNN accuracy as a horizontal dashed line.

In all graphs, the majority of the ODiMO points either dominate the baselines or are part of the Pareto frontier. Importantly, ODiMO produces a \textit{rich set of intermediate Pareto-optimal mappings}, achieving intermediate accuracy vs latency-tradeoffs in-between the various baselines, that could not be obtained otherwise.
This demonstrates the effectiveness of our approach despite the average $1.92\times$ time overhead during training. In fact, while training-time overheads represent a non-recurrent cost, even modest improvements in inference efficiency translate to important gains if scaled up to a real-world large-scale deployment on many devices, and considering the total number of inferences executed during the lifespan of a system.

On DIANA, ODiMO can trade-off the estimated latency and accuracy with respect to the All-8bit baseline on CIFAR-10 (3rd red dot from the right in the top-left graph) achieving a $1.48 \times$ speedup with an accuracy drop lower than $0.5 \%$. Moreover, it also discovers a mapping (1st red dot from the right) that matches the floating-point accuracy, thus improving by +$0.5\%$ the accuracy of All-8bit, thanks to the regularizing effect of ternarization.
On CIFAR-100, our tool discovers solutions spanning more than one order of magnitude on the x-axis, that can offer a speed-up of $1.15\times$ and $4.9\times$ for an accuracy drop $<$$1.5\%$ and $<$$2\%$ w.r.t. the 8bit baseline, respectively (1st and 3rd red dot from the right, top-middle figure). When comparing to the Min-Cost baseline, we can achieve a $0.5\%$ improvement in accuracy at iso-cycles.
Also on the ImageNet task, ODiMO is able to obtain a collection of Pareto points spanning more than one order of magnitude of estimated cycles. In particular, w.r.t. All-8bit we improve the latency by up-to $1.2\times$ and $3\times$ for a drop of $<$$1\%$ and $<$$2.5\%$ accuracy (1st and 2nd red points from the right). 
Moreover, at the cost of $1.08\times$ more cycles, we are able to improve by +$3.8\%$ the Min-Cost accuracy (leftmost red point). This result clearly highlights how ODiMO is able to discover mappings that properly balance accuracy and CUs utilization.

The bottom row of Fig.~\ref{fig:fronts} shows the results obtained targeting the Darkside platform. On the CIFAR-10 benchmark (bottom-left figure) we reduce latency by up to $8 \times$ while being as accurate as the standard convolution baseline (rightmost red point in the bottom-left plot). Instead, when comparing with the Depthwise-Separable baseline (i.e., vanilla MBV1) we can improve accuracy by +$0.7\%$ while also offering a $4.98\times$ speed-up (2nd red point from the right).
On CIFAR-100, ODiMO offers Pareto optimal points spanning two orders of magnitude in latency. Noteworthy, with an accuracy drop respectively of -$1.5\%$ and -$0.1\%$ we achieve in both cases a speed-up of $2.4\times$ when comparing to standard and depthwise-separable convolutions respectively (1st and 4th red points from the right in the bottom-middle plot).
Finally, the rightmost plot in the bottom row of Fig.~\ref{fig:fronts} shows the mappings obtained using ODiMO for Darkside on the ImageNet task. In this case, we limit the search space between the Depthwise and the Depthwise-Separable baselines as corner mappings. To do this, we consider as layer alternatives either DW or DW-Separable (DW + Pointwise) Convolutions, as opposed to DW versus normal Conv. We opted for this choice because, in our preliminary experiments, we discovered that when considering the whole search space the discovered mappings would collapse either on the Depthwise or the Standard Convolution baselines.
Also on this challenging task, we are able to obtain Pareto-optimal mappings. Noteworthy, we can reduce the number of cycles by $1.49\times$ while improving accuracy by $0.2\%$ compared to the Depthwise baseline (leftmost red point).
In the higher accuracy range, the drop with respect to the Depthwise-Separable baseline is 6.8\%, which is traded for a cycles reduction of 2.5$\times$.
\subsubsection{Energy Consumption Optimization}
\begin{figure}[t]
  \centering
  \includegraphics[width=\columnwidth]{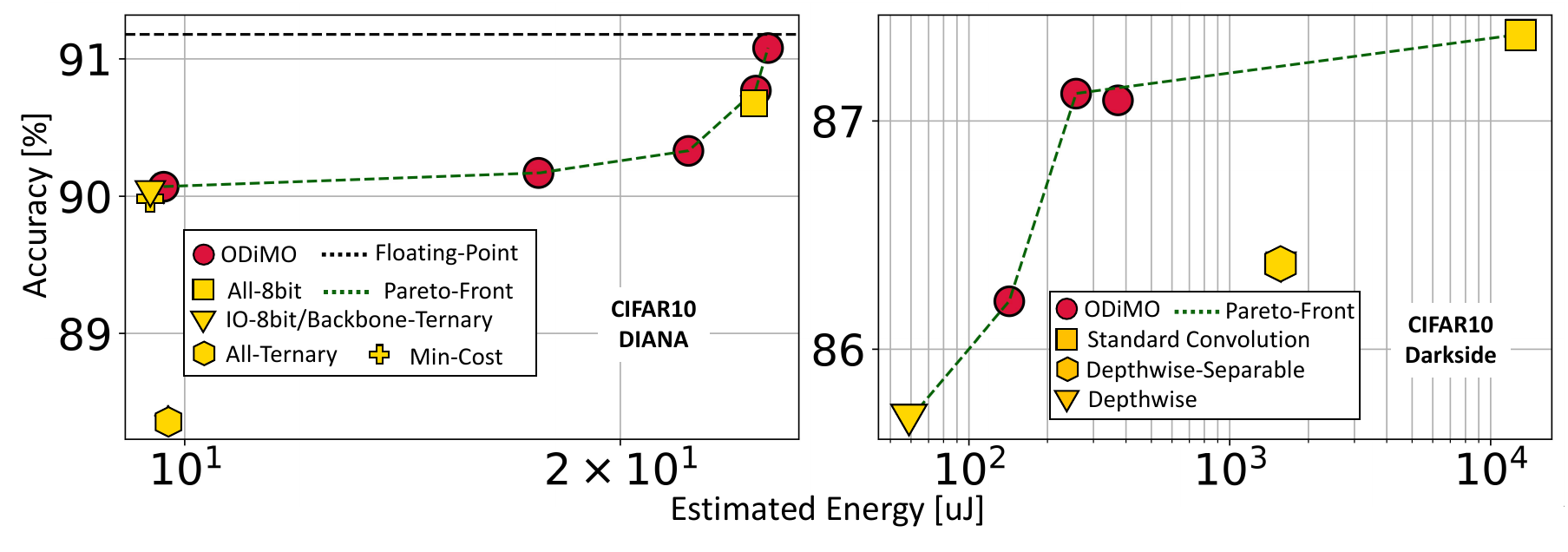}
  \label{fig:test_energy}
  \vspace{-.7cm}
  \caption{Mappings obtained using ODiMO with energy as optimization target.}
\end{figure}
To demonstrate the generality of ODiMO to the specific non-functional cost metric we built an energy cost model in the form of Eq.~\ref{eq:reg_loss_en} for both DIANA and Darkside.
Eq.~\ref{eq:reg_loss_en} allows us to reuse the latency models built for each CU of the two platforms by coupling them with average power consumption numbers. 
In our \href{https://github.com/eml-eda/odimo-journal}{open-source repository} we provide an example of the exact functional form of this optimization target.
Fig.~\ref{fig:test_energy} depicts the results obtained with ODiMO on the CIFAR-10 task. For both hardware platforms, also in this case we obtain Pareto-optimal mappings effectively balancing energy consumption and accuracy.
In particular, in the leftmost plot, we present the results obtained with DIANA. Noteworthy, when accepting an accuracy drop of -$0.37\%$ we can reduce the energy by $1.41\times$ compared to the All-8bit baseline. 
Instead, with the Darkside energy model (rightmost plot), with an accuracy drop of -$0.26\%$ and -$0.15\%$ we can improve the energy efficiency by $50.8\times$ and $11\times$ when comparing respectively with the Standard and Depthwise Convolution baselines. When comparing to the Depthwise-Separable baseline, i.e., the vanilla MobileNet, ODiMO discovered a $6.11\times$ more energy efficient mapping while improving accuracy by $0.76\%$.
\subsection{Additional Comparisons}~\label{subsec:abl_stud}
\begin{figure}[ht]
    \centering
    \includegraphics[width=\linewidth]{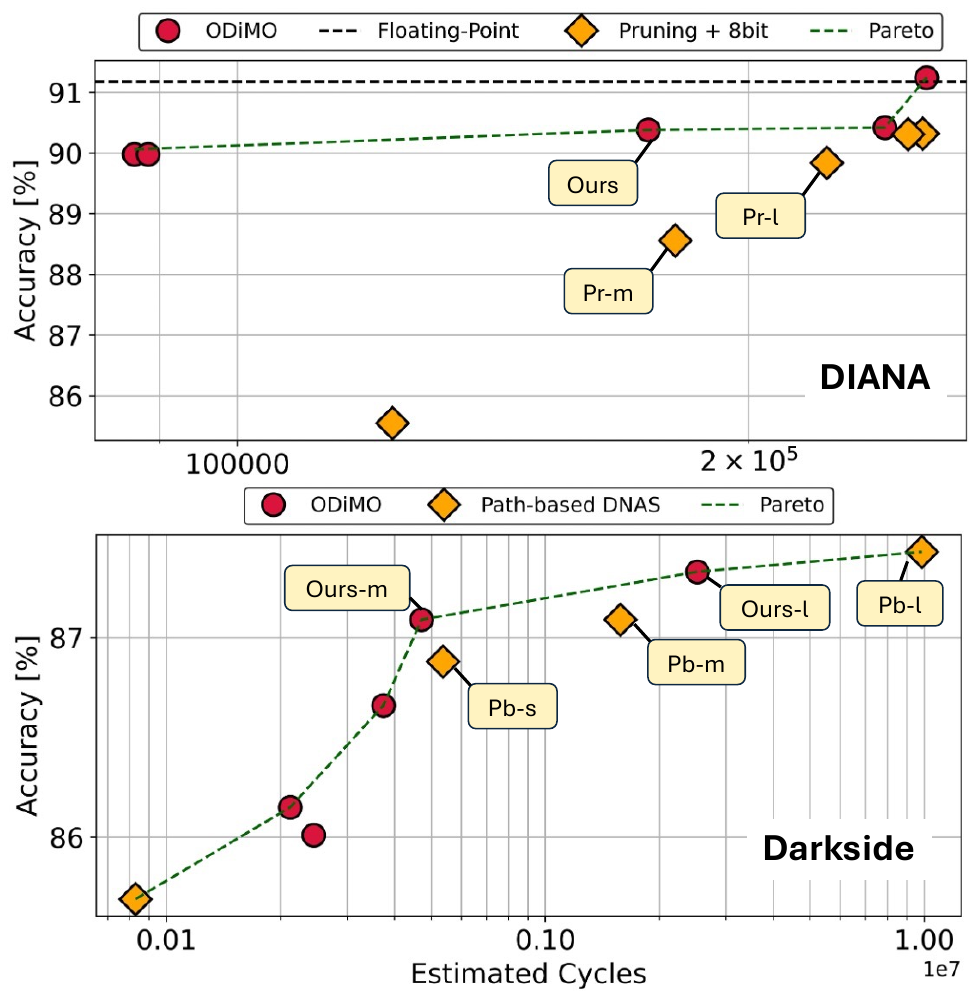}
    \vspace{-.7cm}
    \caption{(Top) Comparison of ODiMO mappings on the DIANA platform for the CIFAR-10 task with structured channel pruning and exclusive mapping on the digital CU; (Bottom) Comparison of ODiMO mappings on the Darkside platform for the CIFAR-10 task with a layer-wise DNAS mapping.}
    \label{fig:pruning_nas_comparisons}
\end{figure}

Fig.~\ref{fig:pruning_nas_comparisons} compares the same ODiMO solutions shown in Fig.~\ref{fig:fronts} for CIFAR-10, with state-of-the-art DNN optimization methods, on both DIANA and Darkside.

In the first plot, our results for DIANA (red circles) are compared with solutions obtained applying a structured channel pruning to the seed network, and then mapping the entire pruned networks, quantized to 8-bit, on the DIANA Digital CU (orange diamonds). In particular, we use the pruning scheme described in~\cite{pit_journal}, as implemented in the open-source PLiNIO~\cite{plinio} library. We prune the same network used as starting point for ODiMO, with different regularization strengths.
All the solutions obtained with pruning followed by full execution on the digital CU are outperformed by ODiMO. In particular, the ODiMO mapping labeled as ``Ours" is more accurate by 0.54\% and faster by 21\% when compared to the pruning solution called ``Pr-l", and more accurate by 1.82\% and faster by 3.6\% when compared to ``Pr-m".
This confirms that mapping the least important channels of each layer to the less precise (but more efficient) Analog CU, can be superior to completely eliminating such channels. 

\begin{figure*}[t]
    \centering
    \includegraphics[width=0.8\textwidth]{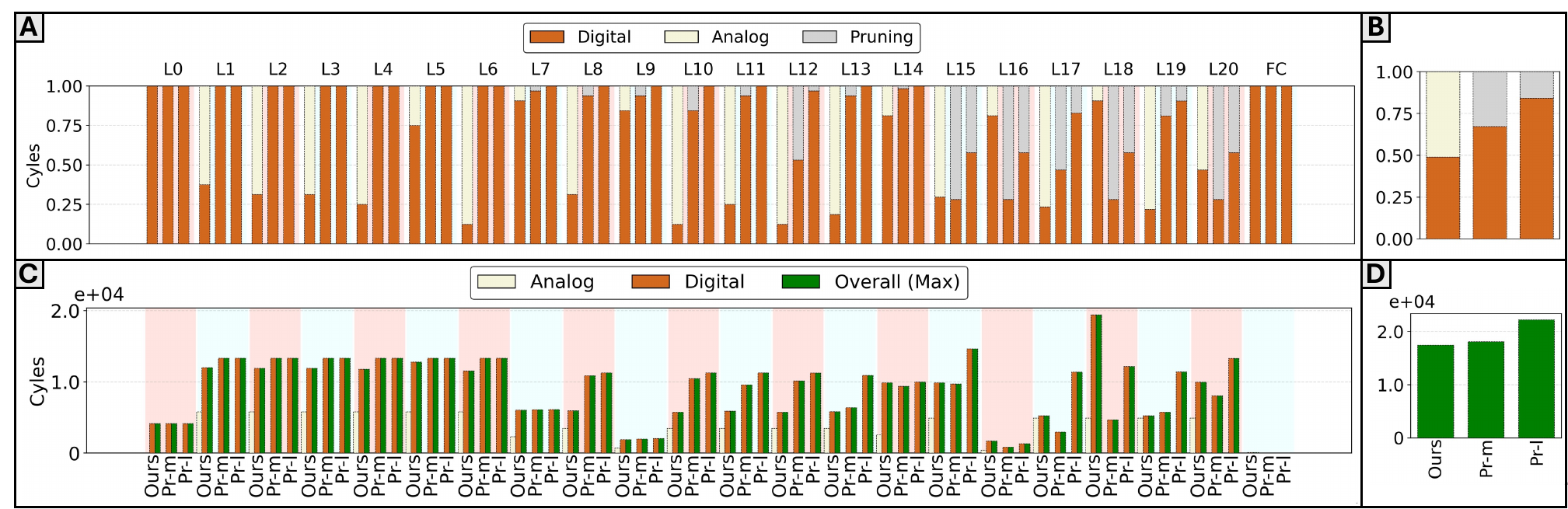}
    \caption{Detailed layer-wise comparison of the ODiMO assignments and latencies for the ``Ours" solution and Pruning ratios for the ``Pr-l" and ``Pr-m" solutions of Fig.~\ref{fig:pruning_nas_comparisons}-Top on the DIANA SoC; (A) Detailed breakdown of layer-wise CU assignment and pruning; (B) Weighted average of CU assignment and pruning across the whole networks; (C) Layer-wise breakdown of the number of cycles executed on each CU; (D) Total number of cycles of each solution.}
    \label{fig:diana_breakdown}
\end{figure*}

Fig.~\ref{fig:diana_breakdown} details this result by showing the fraction of channels mapped to each DIANA CU, or pruned, for all layers of ``Ours'', ``Pr-m'' and ``Pr-l'' as well as the number of estimated cycles executed by each CU in each layer. As it is reasonable, ``Pr-m" has more pruned channels than ``Pr-l". In both cases, pruning is more pronounced in the last layers of the network. 
In contrast, ODiMO is able to reduce the digital channels even in the early layers, by mapping them to the less precise Analog CU, rather than completely eliminating them, thus obtaining a faster yet more accurate solution. 
Moreover, as shown in Fig.~\ref{fig:diana_breakdown}-B the ``Ours" mapping achieves an almost 50\%-50\% balanced usage of both CUs.
Nonetheless, the CU that is bounding the latency is always the digital one, as evident from Fig.~\ref{fig:diana_breakdown}-C, with the analog CU being the fastest in all layers.

\begin{figure*}[t]
    \centering
    \includegraphics[width=0.8\textwidth]{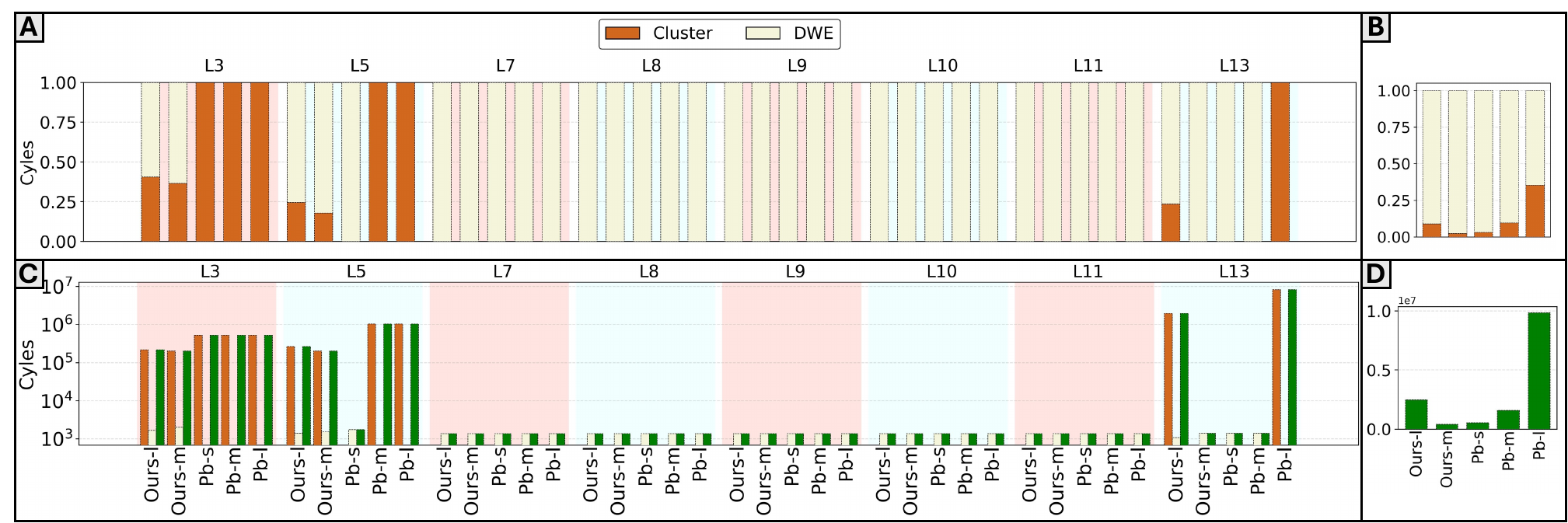}
    \caption{Equivalent of Fig.~\ref{fig:diana_breakdown} for Fig.~\ref{fig:pruning_nas_comparisons}-Bottom on the Darkside SoC.}
    \label{fig:darkside_breakdown}
\end{figure*}

The second plot of Fig.~\ref{fig:pruning_nas_comparisons} compares ODiMO results for Darkside (red circles) to the ones obtained with a DARTS-like~\cite{liu2018darts} path-based DNAS, which coarsely select between alternatives at layer-wise granularity (orange diamonds). Specifically, we build a SuperNet to select between standard and depthwise convolutions for the same layers that we optimize with ODiMO. The layers of the resulting network are then mapped entirely to one of the Darkside's CUs, i.e., the cluster for standard convolutions or DWE for depthwise convolutions.

In the high-cycle region, the layer-wise mapping ``Pb-l" on the Pareto front improves accuracy by 0.1\% over the ODiMO ``Ours-l" solution but requires 3.91$\times$ more cycles. In the mid-cycle range, layer-wise ``Pb-m" and ``Pb-s" are dominated by ODiMO ``Ours-m" which is 3.35$\times$ faster than ``Pb-m" at iso-accuracy, while also 0.21\% more accurate and 1.14$\times$ times faster than ``Pb-s". Moreover, the layer-wise scheme fails to find mappings in the low-cycle region, with all solutions collapsing to mapping all layers on the DWE (one of the heuristic baselines of Fig.~\ref{fig:fronts}).
Fig.~\ref{fig:darkside_breakdown} shows the detailed breakdown of these commented solutions. 
With reference to Fig.~\ref{fig:darkside_breakdown}-A some patterns can be noticed for both mapping schemes: intermediate layers are always mapped to the DWE, while L3 and L5 (near the input), and L13 (near the output) use more flexible standard convolutions on the slower cluster unit. This aligns with the common understanding that input/output-adjacent layers are crucial for accuracy.
However, while in ``Pb-l" and ``Pb-m”, L3 and L5 are fully mapped to the cluster, ``Ours-l" and ``Ours-m,” leveraging ODiMO’s intra-layer mapping capabilities, allocate only part of the channels to the cluster. This balances high accuracy with significantly fewer cycles, as shown in Fig.~\ref{fig:darkside_breakdown}-C and -D. A similar trend is seen in L13, fully mapped to the cluster in ``Pb-l," but only partially in ``Ours-l.”
These results highlight how ODiMO enables new mapping possibilities achieving a good trade-off between accuracy and CUs usage balancing. Indeed, as shown in the breakdowns of Fig.~\ref{fig:darkside_breakdown}, resorting to previous approaches such as layer-wise mapping results in both lower CUs utilization and lower accuracies.
\begin{figure}[t]
  \centering
  \includegraphics[width=0.9\columnwidth]{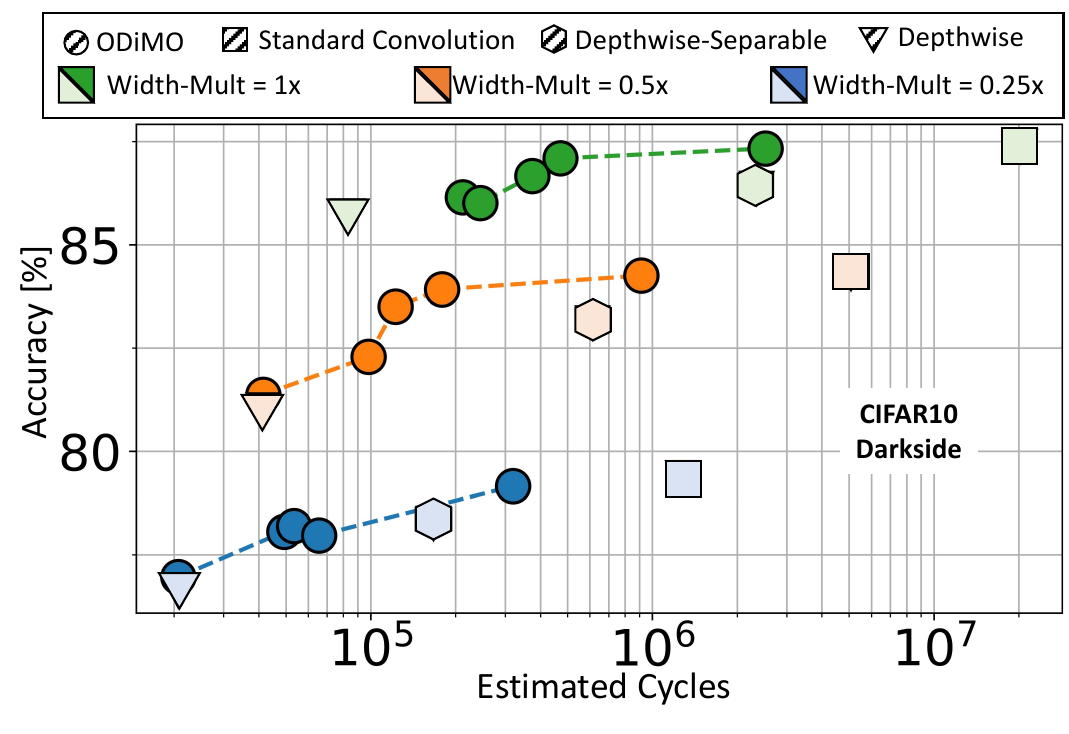}
  \vspace{-.6cm}
  \caption{Mappings obtained using ODiMO with latency as optimization target and different width multiplier.}
  \label{fig:abl_dark}
\end{figure}

In Fig.~\ref{fig:abl_dark} we show the mapping obtained using ODiMO with the Darkside latency model, on the CIFAR-10 task, using the same MBV1 net with three different width-multipliers. Namely, we consider $1\times$ (i.e., the same DNN as in Fig.~\ref{fig:fronts}), $0.5\times$, i.e., with half of the original channels and $0.25\times$, i.e., with a quarter of the original channels.
In all these three variants we always obtain a rich collection of Pareto-optimal mappings.
This demonstrates how ODiMO is effective regardless of the number of channels, which is the geometric dimension of the convolution operation used to perform the partitioning over the different CUs.
Moreover, we can notice how the mappings discovered with a width-multiplier of $1\times$ (green curve) \textit{always} Pareto dominate the solutions obtained with smaller width multipliers (including the baselines). This is due to the high efficiency of the DWE, which can compute a high number of DW output channels with low latency. For instance, when comparing with a vanilla MBV1 with half/quarter of the original channels we are able to improve accuracy by up to $4\%$ and $6.7\%$ at iso-latency.
This experiment is an additional confirmation of the fact that partitioning the execution of each DNN layer over heterogeneous computing units with fine-grained mappings can be a way to reduce latency without incurring the accuracy drop associated with channel pruning.
\subsection{Embedded Deployment}
This section validates the solutions presented in Sec.~\ref{sec:exploration}. We first validate the DIANA and Darkside hardware models with a micro-benchmarking procedure on selected layers. Then, we present the deployment of a subset of the DNNs found by ODiMO on the DIANA SoC. Please note, that while we were able to perform the micro-benchmarking on Darkside, using numbers reported in the paper~\cite{darkside} as ground truth reference, we do not have the possibility of deploying entire networks on it because the physical HW is not available.
\subsubsection{Hardware Models Micro-Benchmarking}~\label{subsec:exp_hw_valid}
\begin{table}[t]
\centering
\caption{Micro-benchmarking of DIANA and Darkside HW models on ResNet and MobileNet layers.}
\vspace{-0.2cm}
\label{tab:microbench}
\begin{tabular}{c|c|c|c|c}
& CU & Error & Pearson & Spearman \\ \cline{1-5}
\multirow{2}{*}{DIANA} & Digital & 42\% & 88\% & 95\% \\
& Analog & 37\% & 79\% & 94\% \\ \cline{1-5}
\multirow{2}{*}{Darkside} & DWE & 9\% & 99.9\% & 99.8\% \\
& Cluster & 16\% & 98\% & 96\% \\ \cline{1-5}
\end{tabular}
\vspace{-0.3cm}
\end{table}
In this section, we validate the four analytical HW models detailed in our \href{https://github.com/eml-eda/odimo-journal}{open-source repository} respectively for DIANA and Darkside. In particular, we compare the number of predicted cycles against the ones measured on the actual SoCs for the same DNN workloads.
The different workloads refer to layers taken from ResNet and MobileNet architectures respectively, with different geometries.
For each model, we computed the Pearson and the Spearman correlation coefficients which measure respectively the strength of the linear relationship and the rank monotonicity between the modeled and real cycles. Moreover, we also computed the average absolute percentage error between the real and the modeled number of cycles.
Table~\ref{tab:microbench} summarizes the results obtained with the four models.
Although some models exhibit relatively high average errors, these are mostly due to neglected latency components, leading to a constant underestimation of the latency. Therefore, the models maintain strong correlations with actual measurements, as indicated by a Spearman coefficient consistently above 94\%, thus being suitable to be used in our NAS algorithm.
In particular, it is important to highlight that the models developed for Darkside show lower errors and improved Pearson/Spearman correlation when compared to the ones of DIANA. This result corroborates the goodness of the ODiMO mappings obtained for the Darkside platform, even if, due to unavailability of the physical chip, we are not able to deploy entire networks on it, as done for DIANA in the next paragraph.
\subsubsection{DIANA Deployment}~\label{subsec:diana_deployment}
This section examines the deployment of a subset of the solutions from Fig.~\ref{fig:fronts} on the DIANA SoC, operating at a frequency of 260 MHz, replacing modeled latency and energy with measured values. For each benchmark, we deploy the All-8bit and Min-Cost baselines along with a selection of ODiMO results (indicated by a black circle in Fig.~\ref{fig:fronts}). Specifically, we select two points from the Pareto front (Accurate and Fast) for all benchmarks.
Each entry of Table~\ref{tab:networkperformance} represents a deployed DNNs for which we report accuracy, latency, energy consumption, the percentage of time each CU is utilized during a complete inference (\textit{D./A. util.}), and the percentage of channels executed on the AIMC CU, i.e., the fraction $\nicefrac{C_{out}^{aimc}}{C_{out}}$ for the whole DNN (\textit{A. Ch.}).
\begin{table}[t]
\caption{Deployment on DIANA of selected solutions from Fig.~\ref{fig:fronts}}
\vspace{-0.2cm}
\label{tab:networkperformance}
\resizebox{\columnwidth}{!}{
\begin{tabular}{l|l|l|l|l|l|l}
                           & Network & Acc. & lat. {[}ms{]}& E. {[}uJ{]} & D./A. util. & A. Ch. \\ \hline
\multirow{4}{*}{CIFAR10} & All-8bit & 90.70 & 1.55 & 38.70 & 100\% / 0\% & 0\% \\
                         & ODiMO Accurate & 91.24 & 1.55 & 42.50 & 100\% / 18.6\% & 5.4\% \\
                         & ODiMO Fast & 90.38 & 1.07 & 34.44 & 100\% / 44.8\% & 51.8\% \\
                         & Min Cost & 90.06 & 0.47 & 13.6 & 9.5\% / 93.6\% & 97.5\% \\ \hline
\multirow{4}{*}{CIFAR100} & All-8bit & 74.10 & 30.3 & 756 & 100\% / 0\% & 0\% \\
                          & ODiMO Accurate & 72.74 & 26.2 & 669
                          & 100\% / 7.3\% & 15\% \\
                          & ODiMO Fast & 71.82 & 2.14 & 65.9 & 70\% / 58\% & 96\% \\
                          & Min Cost & 70.86 & 1.62 & 47.6 & 34\% / 77.3\% & 96\% \\ \hline
\multirow{4}{*}{ImageNet} & All-8bit & 69.33 & 63.2 & 1578 & 100\% / 0\% & 0\% \\
                          & ODiMO Accurate & 66.85 & 33.4 & 881 & 94\% / 31\% & 59\% \\
                          & ODiMO Fast & 62.19 & 4.6 & 136 & 27\% / 87\% & 97\% \\
                          & Min Cost & 58.38 & 4.25 & 129 & 35\% / 86\% & 95\% \\ \hline
\end{tabular}
}
\vspace{-0.3cm}
\end{table}

On CIFAR10, ODiMO-Fast reduces latency by $1.45\times$ w.r.t All-8bit, for a limited accuracy drop (-$0.32\%$). This result validates the $1.48\times$ reduction estimated with the analytical model (ref Sec.~\ref{sec:exploration}). The reduction is achieved by offloading roughly half of the channels to the analog CU.
Further, the digital CU is always active while the AIMC CU is active for $44.8\%$ of the inference time. ODiMO Accurate improves the accuracy of the All-8bit baseline by $0.54\%$ by assigning a small fraction ($5.4\%$) of channels to the analog CU, which probably enforces a regularizing effect.
Moreover, we can appreciate how the ranking of different mappings discovered by ODiMO (i.e., using the analytical models) is preserved in the deployment on the real HW. Indeed, the $1.45\times$ speed-up of ODiMO-Fast w.r.t. ODiMO-Accurate is well-tracked by the DIANA's models, which predict a $1.46\times$ speed-up.

On CIFAR100, ODiMO-Fast reduces latency/energy by $14.2\times$/$11.5\times$ with an accuracy drop $< 2.5\%$ w.r.t. All-8bit. On the other hand, ODiMO Accurate limits the accuracy drop to $1.36\%$ while being $1.16\times$ faster and $1.13\times$ more energy efficient. When comparing ODiMO Fast with the Min Cost baseline, we achieve 0.96\% improved accuracy with a latency/energy overhead of $24.2\%$/$27.8\%$. The fraction of channels assigned to the analog CU is similar for the two mappings (i.e., $96\%$). However, the specific channels assigned to it in different layers change. The accuracy improvement is imputable to this difference, which also causes a higher median usage (+$26\%$) of the digital CU.

On the ImageNet dataset, w.r.t. All-8bit the ODiMO Accurate solution improves latency/energy by $1.89\times$/$1.79\times$ with an accuracy drop $<2.48\%$. This is achieved by offloading $59\%$ of the channels to the analog CU. ODiMO Fast improves the accuracy of Min Cost by $3.81\%$ while being only $1.08\times$/$1.05\times$ slower/less energy efficient. As in the case of CIFAR100, this is achieved by assigning \textit{different} channels to the digital CU w.r.t. the ones selected by the Min Cost heuristic (which is accuracy unaware). Moreover, the latency penalty of such channel assignments to the digital CU is mitigated by an overall higher number of channels assigned to the analog CU ($97\%$ vs $95\%$). This stresses the importance of a fine-grained mapping methodology, as the one proposed in this work, which is both accuracy- and hardware-aware.

\section{Conclusion}
We have introduced ODiMO, a tool that partitions a DNN execution at fine grain among multiple accelerators with incompatible quantization formats or implementing different layer alternatives. To do so, it formulates the problem as a cost-aware differentiable optimization addressed simultaneously with the training of DNN weights. With results on different benchmarks and DNN architectures, we have shown that ODiMO can obtain rich Pareto-fronts in both the accuracy vs energy or latency spaces and reduce latency/energy by up to $8 \times$/$50.8\times$ with limited accuracy drops compared to single-accelerator solutions or heuristic mappings.
As future work directions, we will consider new tasks and DNN architectures coupled with more complex heterogeneous SoCs to demonstrate ODiMO's generalization capability.
Additionally, we will consider more diversified CUs without the strict requirement of having a dedicated shared-memory for activations.
 \label{sec:conclusion}





\ifCLASSOPTIONcaptionsoff
  \newpage
\fi

\bibliographystyle{IEEEtran}
\bibliography{bstctl,references}

\begin{IEEEbiography}[{\includegraphics[width=1in,height=1.25in,clip,keepaspectratio]{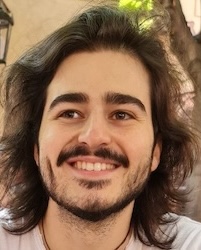}}]{Matteo Risso}
received his B.Sc degree in Physical Engineering and M.Sc degree in Electronic Engineering at the Politecnico di Torino, Italy, in 2018 and 2020. He is currently working toward his Ph.D. degree at Politecnico di Torino, Italy. His research interests include Embedded Machine Learning and Energy-Efficient Embedded Systems.
\end{IEEEbiography}

\begin{IEEEbiography}[{\includegraphics[width=1in,height=1.25in,clip,keepaspectratio]{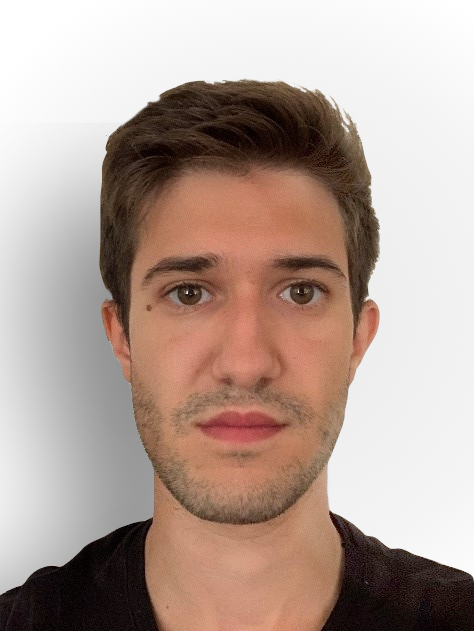}}]{Alessio Burrello}
received his B.Sc and M.Sc degree in Electronic Engineering at the Politecnico of Turin, Italy, in 2016 and 2018.  He is currently working toward his Ph.D. degree at the Department of Electrical, Electronic and Information Technologies Engineering (DEI) of the University of Bologna, Italy.
His research interests include parallel programming models for embedded systems, machine and deep learning, hardware oriented deep learning, and code optimization for multi-core systems.
\end{IEEEbiography}

\begin{IEEEbiography}[{\includegraphics[width=1in,height=1.25in,clip,keepaspectratio]{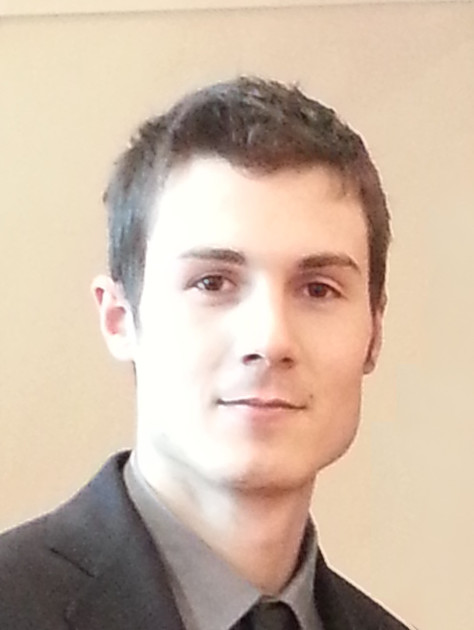}}]{Daniele Jahier Pagliari} received the M.Sc. and Ph.D. degrees in computer engineering from the Politecnico di Torino, Turin, Italy, in 2014 and 2018, respectively. He is currently an Assistant Professor with the Politecnico di Torino. His research interests are in the computer-aided design and optimization of digital circuits and systems, with a particular focus on energy-efficiency aspects and on emerging applications, such as machine learning at the edge.
\end{IEEEbiography}


\end{document}